\documentclass[letterpaper]{article} % DO NOT CHANGE THIS
\usepackage{aaai24}  % DO NOT CHANGE THIS
\usepackage{times}  % DO NOT CHANGE THIS
\usepackage{helvet}  % DO NOT CHANGE THIS
\usepackage{courier}  % DO NOT CHANGE THIS
\usepackage[hyphens]{url}  % DO NOT CHANGE THIS
\usepackage{graphicx} % DO NOT CHANGE THIS
\usepackage{natbib}  % DO NOT CHANGE THIS AND DO NOT ADD ANY OPTIONS TO IT
\usepackage{caption} % DO NOT CHANGE THIS AND DO NOT ADD ANY OPTIONS TO IT
\usepackage{algorithm}
\usepackage{algorithmic}
\usepackage{amsmath}
\usepackage{amssymb}
\usepackage{booktabs}
\usepackage{multirow}
\usepackage{rotating}
\usepackage{bigstrut}
\usepackage{array}

\usepackage{newfloat}
\usepackage{listings}
\DeclareCaptionStyle{ruled}{labelfont=normalfont,labelsep=colon,strut=off} % DO NOT CHANGE THIS
\floatstyle{ruled}
\newfloat{listing}{tb}{lst}{}
\floatname{listing}{Listing}
\nocopyright
\title{DiAD: A Diffusion-based Framework for Multi-class Anomaly Detection}

\author{
    Haoyang He$^1$\thanks{Equal contribution.}
    ~ Jiangning Zhang$^2$\footnotemark[1],
    ~ Hongxu Chen$^1$,
    ~ Xuhai Chen$^1$,
    ~ Zhishan Li$^1$,\\
    ~ \textbf{Xu Chen}$^2$,
    ~ \textbf{Yabiao Wang}$^2$,
    ~ \textbf{Chengjie Wang}$^2$,
    ~ \textbf{Lei Xie}$^1$\thanks{Corresponding author.} \\
    \textnormal{\normalsize $^1$Zhejiang University ~~ $^2$Youtu Lab, Tencent} \\
}

\usepackage{bibentry}
\begin{document}

\maketitle

\begin{abstract}
Reconstruction-based approaches have achieved remarkable outcomes in anomaly detection. The exceptional image reconstruction capabilities of recently popular diffusion models have sparked research efforts to utilize them for enhanced reconstruction of anomalous images. Nonetheless, these methods might face challenges related to the preservation of image categories and pixel-wise structural integrity in the more practical multi-class setting. To solve the above problems, we propose a \textit{\textbf{Di}}fusion-based \textit{\textbf{A}}nomaly \textit{\textbf{D}}etection (\textbf{DiAD}) framework for multi-class anomaly detection, which consists of a pixel-space autoencoder, a latent-space \textit{Semantic-Guided} (SG) network with a connection to the stable diffusion's denoising network, and a feature-space pre-trained feature extractor. Firstly, The SG network is proposed for reconstructing anomalous regions while preserving the original image's semantic information. Secondly, we introduce \textit{Spatial-aware Feature Fusion} (SFF) block to maximize reconstruction accuracy when dealing with extensively reconstructed areas. Thirdly, the input and reconstructed images are processed by a pre-trained feature extractor to generate anomaly maps based on features extracted at different scales. Experiments on MVTec-AD and VisA datasets demonstrate the effectiveness of our approach which surpasses the state-of-the-art methods, \textit{\textbf{e.g.},} achieving 96.8/52.6 and 97.2/99.0 (AUROC/AP) for localization and detection respectively on multi-class MVTec-AD dataset. Code will be available at \url{https://lewandofskee.github.io/projects/diad}.
\end{abstract}

\section{Introduction}
Anomaly detection is a crucial task in computer vision and industrial applications~\cite{9849507, salehi2022unified, liu2023deep}, which goal of visual anomaly detection is to determine anomalous images and locate the regions of anomaly accurately. Existing anomaly detection models~\cite{DBLP:conf/iclr/LiznerskiRVFKM21, Yi_2020_ACCV, yu2021fastflow} mostly correspond to one class, which requires a large amount of storage space and training time as the number of classes increases. Therefore, there is an urgent need for an unsupervised multi-class anomaly detection model that is robust and stable.

\begin{figure}[t]
\centering
\includegraphics[width=0.48\textwidth]{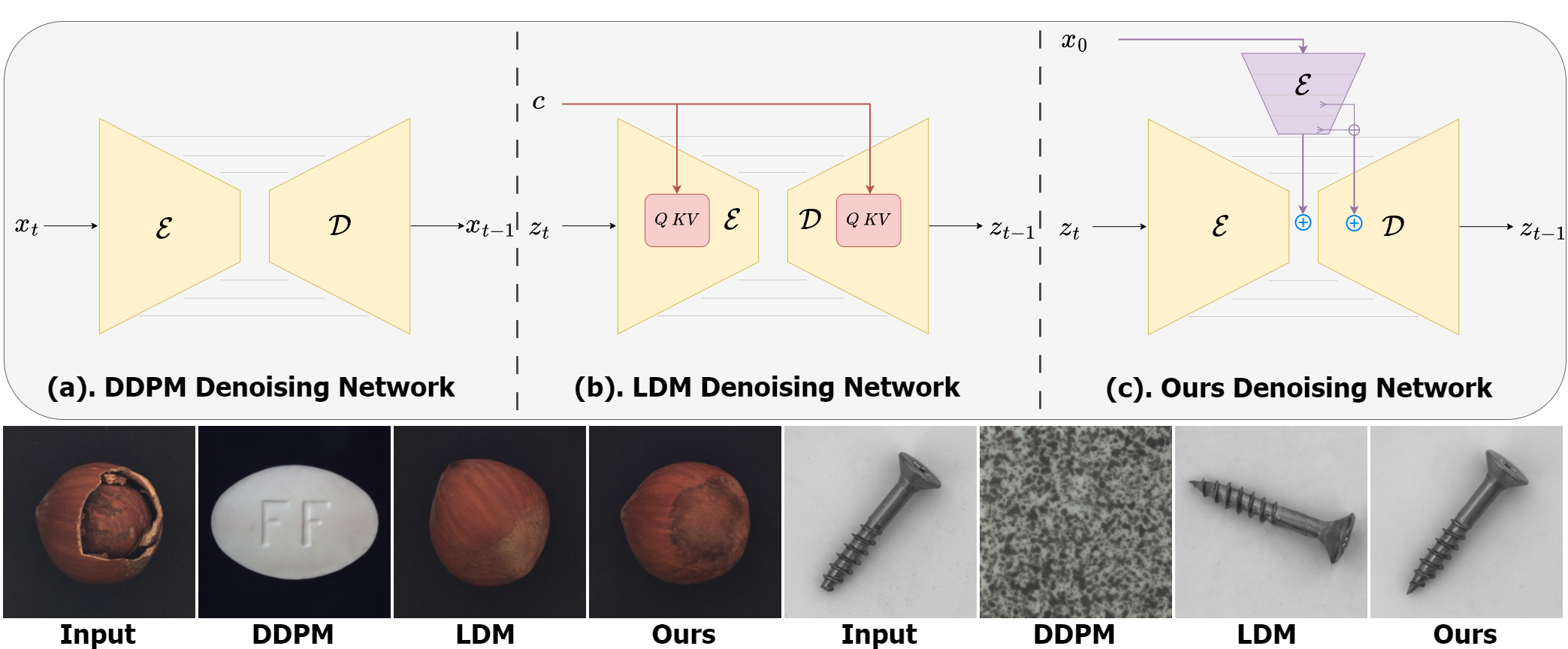} % Reduce the figure size so that it is slightly narrower than the column. Don't use precise values for figure width.This setup will avoid overfull boxes.
\caption{A analysis of different diffusion models for multi-class anomaly detection. The image above shows various denoising network architectures, while the images below demonstrate the results reconstructed by different methods for the same input image. \textbf{\textit{a)}} DDPM suffers from categorical errors. \textbf{\textit{b)}} LDM exhibits semantic errors. \textbf{\textit{c)}} Our approach effectively reconstructs the anomalous regions while preserving the semantic information of the original image.}
\label{motivation}
\end{figure}

The current mainstream unsupervised anomaly detection methods can be divided into three categories: synthesizing-based~\cite{zavrtanik2021draem, li2021cutpaste}, embedding-based~\cite{defard2021padim, roth2022towards, xie2023pushing} and reconstruction-based~\cite{liu2022reconstruction, liang2023omni} methods. The central concept of the reconstruction-based method is that during the training phase, the model only learns from normal images. During the testing phase, the model reconstructs abnormal images into normal ones using the trained model. Therefore, by comparing the reconstructed image with the input image, we can determine the location of anomalies. Traditional reconstruction-based methods, including AEs~\cite{zavrtanik2021reconstruction}, VAEs~\cite{kingma2022autoencoding}, and GANs~\cite{liang2023omni, DBLP:conf/aaai/YanZXHH21} can learn the distribution of normal samples and reconstruct abnormal regions during the testing phase. However, these models have limited reconstruction capabilities and cannot reconstruct complicated textures and objects well, especially large-scale defects or disappearances as shown in Figure~\ref{motivation}. Hence, models with stronger reconstruction capability are required to effectively tackle multi-class anomaly detection.

Recently, the diffusion models~\cite{NEURIPS2020_4c5bcfec, rombach2022highresolution, zhang2023adding} have demonstrated their powerful image-generation capability. However, directly using current mainstream diffusion models cannot effectively address multi-class anomaly detection problems. 1) For the Denoising Diffusion Probabilistic Model (DDPM)~\cite{NEURIPS2020_4c5bcfec} in Fig.~\ref{motivation}-(a), when performing the multi-class setting, this method may encounter issues with misclassifying generated image categories. The reason is that after adding $T$ timesteps noise to the input image, the image has lost its original class information. During inference, denoising is performed based on this Gaussian noise-like distribution, which may generate samples belonging to different categories. 2) Latent Diffusion Model (LDM)~\cite{rombach2022highresolution} has an embedder as a class condition as shown in Fig.~\ref{motivation}-(b), which does not exist the problem of misclassification found in DDPM. However, LDM still cannot address the issue of semantic loss in generated images. LDM is unable to simultaneously preserve the semantic information of the input image while reconstructing the anomalous regions. For example, they may fail to maintain direction consistency with the input image in terms of objects like screws and hazelnuts, as well as exhibit substantial differences from the original image in terms of texture class images.

To address the aforementioned problems, we propose a diffusion-based framework, DiAD, for multi-class anomaly detection and localization, illustrated in Fig.~\ref{arch}, which comprises three components: a pixel space autoencoder, a latent space denoising network and a feature space ImageNet pre-trained model. To effectively maintain consistent semantic information with the original image while reconstructing the location of anomalous regions, we propose the \textbf{S}emantic-\textbf{G}uided (SG) network with a connection to the Stable Diffusion (SD) denoising network in LDM.  To further enhance the capability of preserving fine details in the original image and reconstructing large defects, we propose the \textbf{S}patial-aware \textbf{F}eature \textbf{F}usion (SFF) block to integrate features at different scales. Finally, the reconstructed and input images are passed through a pre-trained model to extract features at different scales and compute anomaly scores. We summarize our contributions as follows:

\begin{itemize}
\item We propose a novel diffusion-based framework DiAD for multi-class anomaly detection, which firstly tackles the problem of existing denoising networks of diffusion-based methods failing to correctly reconstruct anomalies.
\item We construct an SG network connecting to the SD denoising network to maintain consistent semantic information and reconstruct the anomalies.
\item We propose an SFF block to integrate features from different scales to further improve the anomaly reconstruction ability.
\item Abundant experiments demonstrate the sufficient superiority of DiAD over SOTA methods, \textit{\textbf{e.g.},} we surpass the multi-class anomaly detection diffusion-based method by 20.6$\uparrow$/ 11.7$\uparrow$ in pixel/image AUROC and non-diffusion method by 9.2$\uparrow$ in pixel-AP and 0.7$\uparrow$ in image-AUROC on MVTec-AD dataset.
\end{itemize}

\section{Related work}
\subsubsection{Diffusion model.} The diffusion model has gained widespread attention and research interest since its remarkable reconstruction ability. It has demonstrated excellent performance in various applications such as image generation~\cite{zhang2023adding}, video generation~\cite{ho2022imagen}, object detection~\cite{chen2022diffusiondet}, image segmentation~\cite{amit2022segdiff} and etc. LDM~\cite{rombach2022highresolution} introduces conditions through cross-attention to control generation. However, it fails to accurately reconstruct images that contain the original semantic information.

\subsubsection{Anomaly detection.} AD contains a variety of different settings, \textit{e.g.}, open-set~\cite{ding2022catching}, noisy learning~\cite{noisy1, noisy2}, zero-/few-shot~\cite{regad, winclip, saa, aprilgan, clipad, gpt-4v-ad}, 3D AD~\cite{m3dm, easynet}, \textit{etc}. This paper studies general unsupervised anomaly detection, which can primarily be categorized into three major methodologies: 

\textbf{\textit{1)}} Synthesizing-based methods synthesize anomalies on normal image samples. During the training phase, both normal images and synthetically generated abnormal images are input into the network for training, which aids in anomaly detection and localization. DRAEM~\cite{zavrtanik2021draem} consists of an end-to-end network composed of a reconstruction network and a discriminative sub-network, which synthesizes and generates just-out-distribution phenomena. However, due to the diversity and unpredictability of anomalies in real-world scenarios, it is impossible to synthesize all types of anomalies.

\textbf{\textit{2)}} Embedding-based methods encode the original image's three-dimensional information into a multidimensional feature space~\cite{roth2022towards, cao2022informative, memkd}. Most methods employ networks~\cite{resnet, efficientnet, eatformer, emo, pvg} pre-trained on ImageNet~\cite{deng2009imagenet} for feature extraction. RD4AD~\cite{deng2022anomaly} utilizes a WideResNet50~\cite{DBLP:conf/bmvc/ZagoruykoK16} as the teacher model for feature extraction and employs a structurally identical network in reverse as the student model, computing the cosine similarity of corresponding features as anomaly scores. However, due to significant differences between industrial images and the data distribution in ImageNet, the extracted features might not be suitable for industrial anomaly detection purposes.

\textbf{\textit{3)}} Reconstruction-based methods aim to train a model on a dataset without anomalies. The model learns to identify patterns and characteristics in the normal data. OCR-GAN~\cite{liang2023omni} decouples images into different frequencies and uses GAN for reconstruction. EdgRec~\cite{liu2022reconstruction} achieves good reconstruction results by first synthesizing anomalies and then extracting grayscale edge information from images, which is ultimately input into a reconstruction network. However, there are certain limitations in the reconstruction of large-area anomalies. Moreover, the accuracy of anomaly localization is also not sufficient.

\begin{figure*}[t]
\centering
\includegraphics[width=1\textwidth]{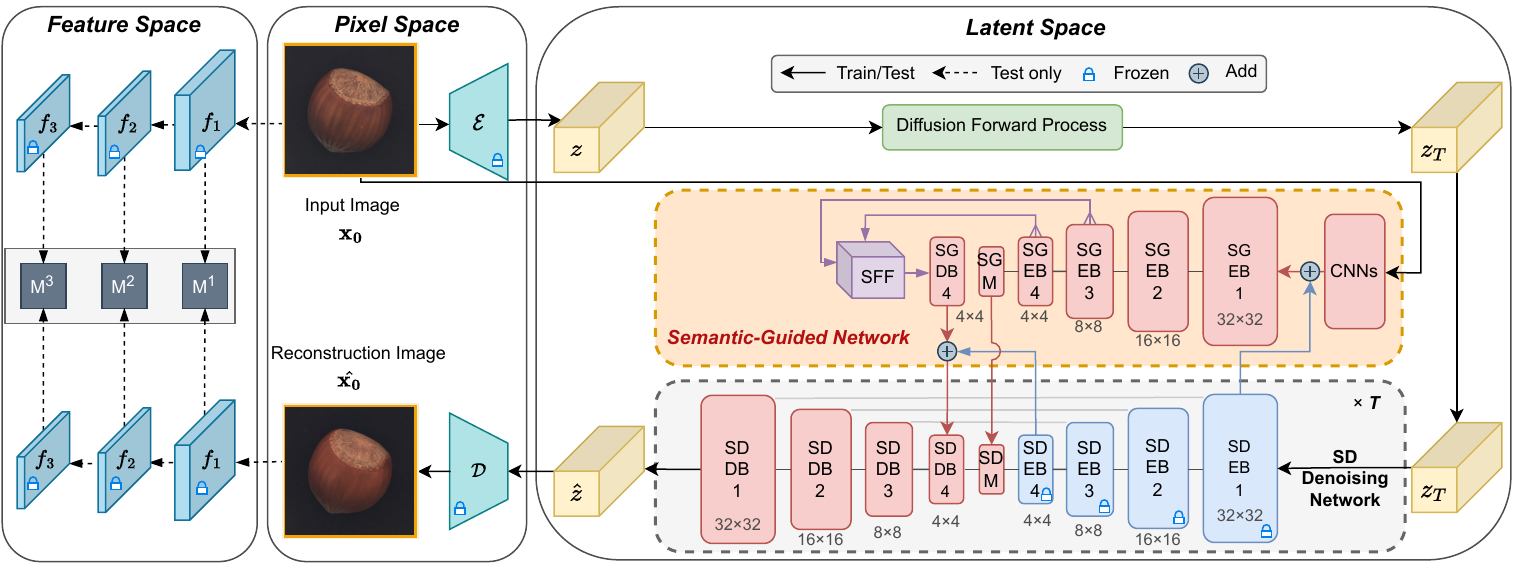} 
\caption{\textbf{Framework of the proposed DiAD that contains three parts}: \textbf{\textit{1)}} a pixel-space autoencoder \{$\mathcal{E},\mathcal{D}$\}; \textbf{\textit{2)}} a latent-space Semantic-Guided (SG) network with a connection to Stable Diffusion (SD) denoising network; and \textbf{\textit{3)}} a feature-space pre-trained feature extractor $\Psi$. During training, the input $x_0$ and the latent variable $z_T$ are inputted into the SG network and the SD denoising network, respectively. The output noise and input noise are calculated for MSE loss and gradient optimization is computed. During testing, $x_0$ and the reconstructed image $\hat{x_0}$ are inputted into the same pre-trained feature extraction network to obtain feature maps \{$f_1$,$f_2$,$f_3$\} of different scales, and their anomaly scores $\mathcal{S}$ are calculated.}
\label{arch}
\end{figure*}

Recently, some studies have applied diffusion models to anomaly detection. AnoDDPM~\cite{DBLP:conf/cvpr/WyattLSW22} is the first approach to employ a diffusion model for medical anomaly detection. DiffusionAD~\cite{zhang2023diffusionad} utilizes an anomaly synthetic strategy to generate anomalous samples and labels, along with two sub-networks dedicated to the tasks of denoising and segmentation. DDAD~\cite{mousakhan2023anomaly} employs a score-based pre-trained diffusion model to generate normal samples while fine-tuning the pre-trained feature extractor to achieve domain transfer. However, these approaches only add limited steps of noise and perform few denoising steps, which makes them unable to reconstruct large-scale defects. 

To overcome the aforementioned problems, We propose a diffusion-based framework DiAD for multi-class anomaly detection, which firstly tackles the problem of existing diffusion-based methods failing to correctly reconstruct anomalies.

\section{Preliminaries}
\subsubsection{Denoising Diffusion Probabilistic Model.} Denoising Diffusion Probabilistic Model (DDPM) consists of two processes: the forward diffusion process and the reverse denoising process. During the forward process, a noisy sample $x_t$ is generated using a Markov chain that incrementally adds Gaussian-distributed noise to an initial data sample $x_0$. The forward diffusion process can be characterized as follows:
\begin{equation}
\label{xt}
x_t=\sqrt{\bar{\alpha}_t} x_0+\sqrt{1-\bar{\alpha}_t} \epsilon_t, \epsilon_t \sim \mathcal{N}(\mathbf{0}, \mathbf{I}),
\end{equation}
where $\alpha_t=1-\beta_t$, $\bar{\alpha}_t=\prod_{i=1}^T \alpha_i=\prod_{i=1}^T (1-\beta_i)$ and $\beta_i$ represents the noise schedule used to regulate the quantity of noise added at each timestep.

In the reverse denoising process, $x_T$ is first sampled from equation~\ref{xt} and $x_{t-1}$ is reconstructed from $x_t$ and the model prediction $\epsilon_\theta\left({x}_t, t\right)$ with the formulation:
\begin{equation}
x_{t-1}=\frac{1}{\sqrt{\alpha_t}}\left(x_t-\frac{1-\alpha_t}{\sqrt{1-\bar{\alpha}_t}} \epsilon_\theta\left(x_t, t\right)\right)+\sigma_t z,
\end{equation}
where $z \sim \mathcal{N}(\mathbf{0}, \mathrm{I})$, $\sigma_t$ is a fixed constant related to the variance schedule, $\epsilon_\theta\left(x_t, t\right)$ is a U-Net~\cite{ronneberger2015u} network to predict the distribution and $\theta$ is the learnable parameter which could be optimized as: 
\begin{equation}
\min _\theta \mathbb{E}_{\boldsymbol{x}_0 \sim q\left(\boldsymbol{x}_0\right), \epsilon \sim \mathcal{N}(\mathbf{0}, \mathbf{I}), t}\left\|\epsilon-\epsilon_\theta\left(\boldsymbol{x}_t, t\right)\right\|_2^2.
\end{equation}

\subsubsection{Latent Diffusion Model.} 
Latent Diffusion Model (LDM) focuses on the low-dimensional latent space with conditioning mechanisms. LDM consists of a pre-trained autoencoder model and a denoising U-Net-like attention-based network. The network compresses images using an encoder, conducts diffusion and denoising operations in the latent representation space, and subsequently reconstructs the images back to the original pixel space using a decoder. The training optimization objective is:
\begin{equation}
\mathcal{L}_{L D M}=\mathbb{E}_{z_0, t, c, \epsilon \sim \mathcal{N}(0,1)}\left[\left\|\epsilon-\epsilon_\theta\left(z_t, t, c\right)\right\|_2^2\right],
\end{equation}
where $c$ represents the conditioning mechanisms which can consist of multimodal types such as text or image, connected to the model through a cross-attention mechanism. $z_t$ represents the latent space variable, 
% while $\epsilon_\theta\left(x_t, t\right)$ signifies a U-Net model with attention-based architecture. 

\section{Method}
The proposed pipeline DiAD is shown in Fig.~\ref{arch}. First, the pre-trained encoder downsamples the input image into a latent-space representation. Then, noise is added to the latent representation, followed by the denoising process using an SD denoising network with a connection to the SG network. The denoising process is repeated for the same timesteps as the diffusion process. Finally, the reconstructed latent representation is restored to the original image level using the pre-trained decoder. In terms of anomaly detection and localization, the input and reconstructed images are fed into the same pre-trained model to extract features at different scales and calculate the differences between these features.

\subsection{Semantic-Guided Network}
As discussed earlier, DDPM and LDM each have specific problems when addressing multi-class anomaly detection tasks. In response to these issues and the multi-class task itself, we propose an SG network to address the problem of LDM's inability to effectively reconstruct anomalies and preserve the semantic information of the input image.

Given an input image $x_0 \in \mathbb{R}^{3\times H\times W}$ in pixel space, the pre-trained encoder $\mathcal{E}$ encodes $x_0$ into a latent space representation $z = \mathcal{E}(x_0)$ where $z \in \mathbb{R}^{c\times h\times w}$. Similar to Eq.~\ref{xt} where the original pixel space $x$ is replaced by latent representation $z$, the forward diffusion process now can be characterized as follows:
\begin{equation}
z_t=\sqrt{\bar{\alpha}_t} z_0+\sqrt{1-\bar{\alpha}_t} \epsilon_t, \epsilon_t \sim \mathcal{N}(\mathbf{0}, \mathbf{I}).
\end{equation}

The perturbed representation $z_T$ and input $x_0$ are simultaneously fed into the SD denoising network and SG network, respectively. After $T$ steps of the reverse denoising process, the final variable $\hat{z}$ is restored to the reconstructed image $\hat{x_0}$ from the pre-trained decoder $\mathcal{D}$ giving $\hat{x_0} = \mathcal{D}(\hat{z})$. The training objective of DiAD is:
\begin{equation}
\mathcal{L}_{DiAD}=\mathbb{E}_{z_0, t, c_i, \epsilon \sim \mathcal{N}(0,1)}\left[\left\|\epsilon-\epsilon_\theta\left(z_t, t, c_i\right)\right\|_2^2\right].
\end{equation}

The denoising network consists of a pre-trained SD denoising network and an SG network that replicates the SD parameters for initiation as shown in Fig.~\ref{arch}. The pre-trained SD denoising network comprises four encoder blocks, one middle block and four decoder blocks. Here, 'block' means a frequently utilized unit in the construction of the neural network layer, \textit{\textbf{e.g.},}, 'resnet' block, transformer block, multi-head cross attention block, \textit{etc}.

The input image $x_0 \in \mathbb{R}^{3\times H\times W}$ is transformed into $\mathrm{x} \in \mathbb{R}^{d\times h\times w}$ by a set of 'conv-silu' layers $\mathcal{C}$ in SG network in order to keep the same dimension with the latent representations in SD Encoder Block 1 $\mathcal{E}_{SD1}$. Then, the result of the summation of $\mathrm{x}$ and $z$ are input into the SG Encoder Blocks (SGEBs). After continuous downsampling by the encoder $\mathcal{E}_{SG}$, the results are finally added to the output of the SD middle block $\mathcal{M}_{SD}$ after its completion in the middle block $\mathcal{M}_{SG}$. Additionally, to address multi-class tasks of different scenarios and categories, the results of the SG Decoder Blocks (SGDBs) $\mathcal{D}_{SG}$ are also added to the results of the SD decoder $\mathcal{D}_{SD}$ with an SFF block combined which will be particularly explained in the next section. The output $\mathcal{G}$ of the denoising network is characterized as:
\begin{equation}
\begin{aligned}
\label{fea2}
\mathcal{G} =\mathcal{D}_{S D} & \left(\mathcal{M}_{S D}\left(\mathcal{E}_{S D}\left(z_t\right)\right)+\right. \left.\mathcal{M}_{SG}\left(\mathcal{E}_{S D}\left(z+ \mathcal{C}\left(x_0\right)\right)\right)\right)\\
&\ + \mathcal{D}_{SG j}(\mathcal{M}_{SG}\left(\mathcal{E}_{S D}\left(z+ \mathcal{C}\left(x_0\right)\right)\right)),
\end{aligned}
\end{equation}
where $z$ represents the latent representation with noise perturbed, $x_0$ represents the input image, $\mathcal{C}(\cdot)$ represents a set of 'conv-silu' layers in SG network, $\mathcal{E}_{S D}(\cdot)$ represents all the SD encoder blocks (SDEBs), $\mathcal{E}_{SG}(\cdot)$ represents all the SGEBs, $\mathcal{M}_{SG}(\cdot)$ and $\mathcal{M}_{SD}(\cdot)$ represent SG and SD middle blocks respectively, $\mathcal{D}_{SD}(\cdot)$ represent all the SDDBs and $\mathcal{D}_{SG j}(\cdot)$ represents SGDBs for $j$-th blocks.

\begin{table*}[htbp]
  \centering
  \resizebox{1.0\linewidth}{!}{
    \begin{tabular}{p{3em}<{\centering} p{3.25em}<{\centering}|p{3em}<{\centering} p{3em}<{\centering} p{6.2em}<{\centering} p{6.2em}<{\centering} p{6.2em}<{\centering} |p{6.2em}<{\centering} p{6.2em}<{\centering} | p{7em}<{\centering}}
    \toprule
    \multicolumn{2}{c|}{\multirow{2}[4]{*}{Category}} & \multicolumn{5}{c|}{Non-Diffusion Method} & \multicolumn{3}{c}{Diffusion-based Method} \\
\cmidrule{3-10}    \multicolumn{2}{c|}{} & \multicolumn{1}{c}{PaDiM} & \multicolumn{1}{c}{MKD} & \multicolumn{1}{c}{DRAEM} & \multicolumn{1}{c}{RD4AD} & \multicolumn{1}{c|}{UniAD} & DDPM  & LDM   & Ours \\
    \midrule
    \multicolumn{1}{c|}{\multirow{10}[1]{*}{\begin{turn}{-90}Objects\end{turn}}} & \multicolumn{1}{c|}{Bottle} & 97.9/- & 98.7/- & 97.5/99.2/96.1 & 99.6/99.9/98.4 & \textbf{99.7}/\textbf{100.}/\textbf{100.} & 63.6/71.8/86.3 & 93.8/98.7/93.7 & 99.7/96.5/91.8 \\
    \multicolumn{1}{c|}{} & \multicolumn{1}{c|}{Cable} & 70.9/- & 78.2/- & 57.8/74.0/76.3 & 84.1/89.5/82.5 & \textbf{95.2}/95.9/88.0 & 55.6/69.7/76.0 & 55.7/74.8/77.7 & 94.8/\textbf{98.8}/\textbf{95.2} \\
    \multicolumn{1}{c|}{} & \multicolumn{1}{c|}{Capsule} & 73.4/- & 68.3/- & 65.3/92.5/90.4 & \textbf{94.1}/96.9/\textbf{96.9} & 86.9/\textbf{97.8}/94.4 & 52.9/82.0/90.5 & 60.5/81.4/90.5 & 89.0/97.5/95.5 \\
    \multicolumn{1}{c|}{} & \multicolumn{1}{c|}{Hazelnut} & 85.5/- & 97.1/- & 93.7/97.5/92.3 & 60.8/69.8/86.4 & \textbf{99.8}/\textbf{100.}/\textbf{99.3} & 87.0/90.4/88.1 & 93.0/95.8/89.8 & 99.5/99.7/97.3 \\
    \multicolumn{1}{c|}{} & \multicolumn{1}{c|}{Metal Nut} & 88.0/- & 64.9/- & 72.8/95.0/92.0 & \textbf{100.}/\textbf{100.}/\textbf{99.5} & 99.2/99.9/\textbf{99.5} & 60.0/74.4/89.4 & 53.0/80.1/89.4 & 99.1/96.0/91.6 \\
    \multicolumn{1}{c|}{} & \multicolumn{1}{c|}{Pill} & 68.8/- & 79.7/- & 82.2/94.9/92.4 & \textbf{97.5}/\textbf{99.6}/\textbf{96.8} & 93.7/98.7/95.7 & 55.8/84.0/91.6 & 62.1/93.1/91.6 & 95.7/98.5/94.5 \\
    \multicolumn{1}{c|}{} & \multicolumn{1}{c|}{Screw} & 56.9/- & 75.6/- & 92.0/95.7/89.9 & \textbf{97.7}/99.3/95.8 & 87.5/96.5/89.0 & 53.6/71.9/85.9 & 58.7/81.9/85.6 & 90.7/\textbf{99.7}/\textbf{97.9} \\
    \multicolumn{1}{c|}{} & \multicolumn{1}{c|}{Toothbrush} & 95.3/- & 75.3/- & 90.6/96.8/90.0 & 97.2/99.0/94.7 & 94.2/97.4/95.2 & 57.5/68.0/83.3 & 78.6/83.9/83.3 & \textbf{99.7}/\textbf{99.9}/\textbf{99.2} \\
    \multicolumn{1}{c|}{} & \multicolumn{1}{c|}{Transistor} & 86.6/- & 73.4/- & 74.8/77.4/71.1 & 94.2/95.2/90.0 & 99.8/98.0/93.8 & 57.8/44.6/57.1 & 61.0/57.8/59.1 & \textbf{99.8}/\textbf{99.6}/\textbf{97.4} \\
    \multicolumn{1}{c|}{} & \multicolumn{1}{c|}{Zipper} & 79.7/- &  87.4/- & 98.8/\textbf{99.9}/99.2 & \textbf{99.5}/\textbf{99.9}/\textbf{99.2} & 95.8/99.5/97.1 & 64.9/77.4/88.1 & 73.6/89.5/90.6 & 95.1/99.1/94.4 \\
    \midrule
    \multicolumn{1}{c|}{\multirow{5}[1]{*}{\begin{turn}{-90}Textures\end{turn}}} & \multicolumn{1}{c|}{Carpet} & 93.8/- & 69.8/- & 98.0/99.1/96.7 & 98.5/99.6/97.2 & \textbf{99.8}/\textbf{99.9}/\textbf{99.4} & 95.5/98.7/91.0 & 99.4/99.8/\textbf{99.4} & 99.4/\textbf{99.9}/98.3 \\
    \multicolumn{1}{c|}{} & \multicolumn{1}{c|}{Grid} & 73.9/- & 83.8/- & \textbf{99.3}/99.7/\textbf{98.2} & 98.0/99.4/96.5 & 98.2/99.5/97.3 & 83.5/93.9/86.9 & 67.3/82.6/84.4 & 98.5/\textbf{99.8}/97.7 \\
    \multicolumn{1}{c|}{} & \multicolumn{1}{c|}{Leather} & 99.9/- & 93.6/- & 98.7/99.3/95.0 & \textbf{100.}/\textbf{100.}/\textbf{100.} & \textbf{100.}/\textbf{100.}/\textbf{100.} & 98.4/99.5/96.3 & 97.4/99.0/96.3 & 99.8/99.7/97.6 \\
    \multicolumn{1}{c|}{} & \multicolumn{1}{c|}{Tile} & 93.3/- & 89.5/- & \textbf{99.8}/\textbf{100.}/\textbf{100.} & 98.3/99.3/96.4 & 99.3/99.8/98.2 & 93.697.5/92.0 & 97.1/98.7/94.1 & 96.8/99.9/98.4 \\
    \multicolumn{1}{c|}{} & \multicolumn{1}{c|}{Wood} & 98.4/- & 93.4/- & \textbf{99.8}/\textbf{100.}/\textbf{100.} & 99.2/99.8/98.3 & 98.6/99.6/96.6 & 98.6/99.6/97.5 & 97.8/99.4/95.9 & 99.7/\textbf{100.}/\textbf{100.} \\
    \midrule
    \multicolumn{2}{c|}{Mean} & 84.2/- & 81.9/- & 88.1/94.7/92.0 & 94.6/96.5/95.2 & 96.5/98.8/96.2 & 71.9/81.6/86.6 & 76.6/87.8/88.1 & \textbf{97.2}/\textbf{99.0}/\textbf{96.5} \\
    \bottomrule
    \end{tabular}%
  }
  \caption{Comparison with SOTA methods on MVTec-AD dataset for multi-class anomaly detection with $AUROC_{cls}$/$AP_{cls}$/$F1max_{cls}$ metrics.}
  \label{tab:mvtecsp}%
\end{table*}%

\subsection{Spatial-aware Feature Fusion Block}
\begin{figure}[t]
\centering
\includegraphics[width=0.45\textwidth]{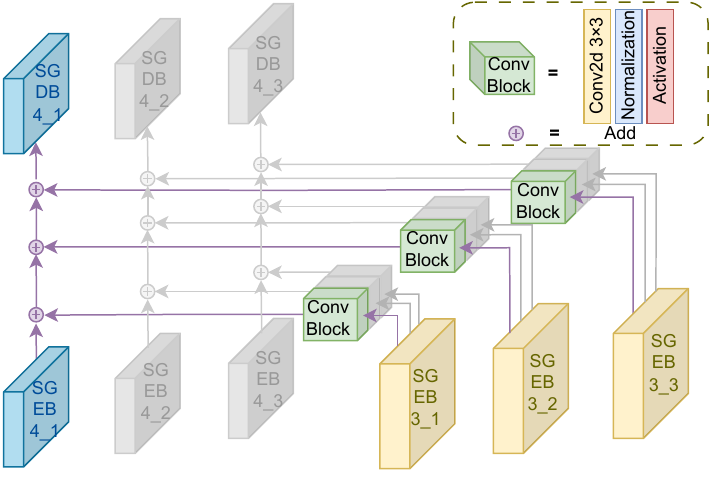}
\caption{\textbf{Schematic diagram of SFF block}. Each layer in SGDB4 is obtained by adding the corresponding SGEB4 to every SGEB3 with Conv Block performed.} 
\label{sff}
\end{figure}
When adding several layers of decoder blocks from SGEBs to SDDBs during the experiment as shown in Table~\ref{tab:diad}, we found it to be challenging to solve the multi-class anomaly detection. This is because the dataset contains various types, such as objects and textures. For texture-related cases, the anomalies are generally smaller, so it is necessary to preserve their original textures. On the other hand, the defects often cover larger areas for object-related cases, requiring stronger reconstruction capabilities. Therefore, it is extremely challenging to simultaneously preserve the normal information of the original samples and reconstruct the abnormal locations in different scenarios.

Hence, we proposed a Spatial-aware Feature Fusion (SFF) block with the aim of integrating high-scale semantic information into the low-scale. This ultimately enables the model to both preserve the information of the original normal samples and reconstruct large-scale abnormal regions. The structure of the SFF block is shown in Fig.~\ref{sff}. Each SGEBs consists of three sub-layers. Therefore, the SFF block integrates the features of each layer in SGEB3 into each layer in SGEB4 and adds the fused features to the original features. The final output of each layer of the SGEB4 is:

\begin{equation}
    \mathcal{Q}_i = \mathcal{P}_i + \sum_{j=1}^{J} \mathcal{F}(\mathcal{H}_j),
\end{equation}
where $\mathcal{P}_i$ represents the low-scale output features of the $i$-th layer of SGEB4, $\mathcal{Q}_i$ represents the final low-scale output features of the $i$-th layer of SGDB4, $\mathcal{H}_j$ represents the high-scale output features of the $j$-th layer of SGEB3, $J=3$ indicates three layers of SGEB3 used in the experiment and $\mathcal{F}(\cdot)$ represent a basic convolutional block which consists of a 3x3 convolution layer followed by a normalization layer and an activation layers.

As Batch Normalization (BN)~\cite{DBLP:conf/icml/IoffeS15} considers the normalization statistics of all images within a batch, it leads to a loss of unique details in each sample. BN is suitable for a relatively large mini-batch scenario with similar data distributions. However, for multi-class anomaly detection where there are significant differences in data distributions among different categories, normalizing the entire batch is not suitable for tasks in the multi-class setting. Since the results generated by using SD mainly depend on the input image instance, using Instance Normalization (IN)~\cite{ulyanov2017instance} can not only accelerate model convergence but also maintain the independence between each image instance. In addition, in terms of choosing the activation function, we use the SiLU~\cite{elfwing2018sigmoid} instead of the commonly used ReLU~\cite{hahnloser2000digital}, which can preserve more input information. Experimental results in Table~\ref{tab:diad} show that the performance is improved by using IN and SiLU simultaneously instead of the combination of BN and ReLU.

\subsection{Anomaly localization and detection}
During the inference stage, the reconstruction image is obtained through the diffusion and denoising process in the latent space. For anomaly localization and detection, We use the same ImageNet pre-trained feature extractor $\Psi$ to extract features from both the input image $x_0$ and the reconstructed image $\hat{x_0}$ and calculate the anomaly map on different scale feature maps $\mathcal{M}^n$ using cosine similarity:

\begin{equation}
\mathcal{M}^n(x_0, \hat{x_0})=1-\frac{\left(\Psi^n(x_0, \hat{x_0})\right)^T \cdot \Psi^n(x_0, \hat{x_0})}{\left\|\Psi^n(x_0, \hat{x_0})\right\|\left\|\Psi^n(x_0, \hat{x_0})\right\|},
\end{equation}
where $n$ represents the $n$-th feature layer $f_n$ and the anomaly score $\mathcal{S}$ for an input-pair of anomaly localization is:

\begin{equation}
\mathcal{S} = \sum_{n\in N}{\sigma_n \mathcal{M}^n(x_0, \hat{x_0})},
\end{equation}

\begin{table}[htbp]
  \centering
  \resizebox{\linewidth}{!}{
    \begin{tabular}{p{7em}<{\centering}|p{3em}<{\centering}p{3em}<{\centering}|p{3em}<{\centering}p{3em}<{\centering}|p{4em}<{\centering}}
    \toprule
    \multicolumn{1}{c|}{\multirow{2}[3]{*}{Metrics}} & \multicolumn{2}{c|}{Non-Diffusion} & \multicolumn{3}{c}{Diffusion-based} \\
\cmidrule{2-6}    \multicolumn{1}{c|}{} & DRAEM & UniAD & DDPM  & LDM   & Ours \\
    \midrule
    $AUROC_{cls}$ & 79.1 & 85.5 & 54.5 & 56.7 & \textbf{86.8} \\
    $AP_{cls}$ & \multicolumn{1}{c}{81.9} & \multicolumn{1}{c|}{85.5} & \multicolumn{1}{c}{57.9} & \multicolumn{1}{c|}{61.4} & \multicolumn{1}{c}{\textbf{88.3}} \\
    $F1max_{cls}$ & \multicolumn{1}{c}{78.9} & \multicolumn{1}{c|}{84.4} & \multicolumn{1}{c}{72.3} & \multicolumn{1}{c|}{73.1} & \multicolumn{1}{c}{\textbf{85.1}} \\
    \midrule
    $AUROC_{seg}$ & \multicolumn{1}{c}{91.3} & \multicolumn{1}{c|}{95.9} & \multicolumn{1}{c}{79.7} & \multicolumn{1}{c|}{86.6} & \textbf{96.0} \\
    $AP_{seg}$ & \multicolumn{1}{c}{23.5} & \multicolumn{1}{c|}{21.0} &  \;\;2.2   & \;\;6.0     & \textbf{26.1} \\
    $F1max_{seg}$ & 29.5  & 27.0    & \;\;4.5   & \;\;9.9   & \textbf{33.0} \\
    $PRO$ & 58.8  & \textbf{75.6}  & 46.8  & 55.0    & 75.2 \\
    \bottomrule
    \end{tabular}}
    \caption{Quantitative comparisons on VisA dataset.}
  \label{tab:visa}%
\end{table}%

where $\sigma_n$ indicates the upsampling factor in order to keep the same dimension of the pixel space image and $N$ indicates the number of feature layers used during inference.

\section{Experiment}
\subsection{Datasets and evaluation metrics}
\noindent\textbf{MVTec-AD dataset.} MVTec-AD~\cite{bergmann2019mvtec} dataset simulates real-world industrial production scenarios, filling the gap in unsupervised anomaly detection. It consists of 5 types of textures and 10 types of objects, in 5,354 high-resolution images from different domains. The training set contains 3,629 images with only anomaly-free samples. The test set consists of 1,725 images, including both normal and abnormal samples. Pixel-level annotations are provided for the anomaly localization evaluation.

\noindent\textbf{VisA dataset.} VisA~\cite{zou2022spot} dataset consists of a total of 10,821 high-resolution images, including 9,621 normal images and 1,200 anomaly images with 78 types of anomalies. The VisA dataset comprises 12 subsets, each corresponding to a distinct object. 12 objects could be categorized into three different object types: Complex structure, Multiple instances, and Single instance.

\begin{table*}[htbp]
  \centering
  \resizebox{1.0\linewidth}{!}{
    \begin{tabular}{p{3em}<{\centering} p{3.25em}<{\centering}|p{3em}<{\centering} p{3em}<{\centering} p{6.2em}<{\centering} p{6.2em}<{\centering} p{6.2em}<{\centering} |p{6.2em}<{\centering} p{6.2em}<{\centering} | p{7em}<{\centering}}
    \toprule
    \multicolumn{2}{c|}{\multirow{2}[4]{*}{Category}} & \multicolumn{5}{c|}{Non-Diffusion Method} & \multicolumn{3}{c}{Diffusion-based Method} \\
\cmidrule{3-10}    \multicolumn{2}{c|}{} & \multicolumn{1}{c}{PaDiM} & \multicolumn{1}{c}{MKD} & \multicolumn{1}{c}{DRAEM} & \multicolumn{1}{c}{RD4AD} & \multicolumn{1}{c|}{UniAD} & DDPM  & LDM   & Ours \\
    \midrule
    \multicolumn{1}{c|}{\multirow{10}[1]{*}{\begin{turn}{-90}Objects\end{turn}}} & \multicolumn{1}{c|}{Bottle} & 96.1/- & 91.8/- & 87.6/62.5/56.9 & 97.8/\textbf{68.2}/67.6 & 98.1/66.0/\textbf{69.2} & 59.9/ \;4.9/11.7 & 86.9/49.1/50.0 & \textbf{98.4}/52.2/54.8 \\
    \multicolumn{1}{c|}{} & \multicolumn{1}{c|}{Cable} & 81.0/- & 89.3/- & 71.3/14.7/17.8 & 85.1/26.3/33.6 & 97.3/39.9/45.2 & 66.5/ \;6.7/10.6 & 89.3/18.5/26.2 & \textbf{96.8}/\textbf{50.1}/\textbf{57.8} \\
    \multicolumn{1}{c|}{} & \multicolumn{1}{c|}{Capsule} & 96.9/- & 88.3/- & 50.5/ \;6.0/10.0 & \textbf{98.8}/\textbf{43.4}/\textbf{50.0} & 98.5/42.7/46.5 & 63.1/ \;6.2/ \;9.7 & 90.0/ \;7.9/27.3 & 97.1/42.0/45.3 \\
    \multicolumn{1}{c|}{} & \multicolumn{1}{c|}{Hazelnut} & 96.3/- & 91.2/- & 96.9/70.0/60.5 & 97.9/36.2/51.6 & 98.1/55.2/56.8 & 91.2/24.1/28.3 & 95.1/51.2/53.5 & \textbf{98.3}/\textbf{79.2}/\textbf{80.4} \\
    \multicolumn{1}{c|}{} & \multicolumn{1}{c|}{Metal Nut} & 84.8/- & 64.2/- & 62.2/31.1/21.0 & 93.8/\textbf{62.3}/65.4 & 94.8/55.5/\textbf{66.4} & 62.7/14.6/29.2 & 70.5/19.3/30.7 & \textbf{97.3}/30.0/38.3 \\
    \multicolumn{1}{c|}{} & \multicolumn{1}{c|}{Pill} & 87.7/- & 69.7/- & 94.4/59.1/44.1 & \textbf{97.5}/\textbf{63.4}/\textbf{65.2} & 95.0/44.0/53.9 & 55.3/ \;4.0/ \;8.4 & 74.9/10.2/15.0 & 95.7/46.0/51.4 \\
    \multicolumn{1}{c|}{} & \multicolumn{1}{c|}{Screw} & 94.1/- & 92.1/- & 95.5/33.8/40.6 & \textbf{99.4}/40.2/44.6 & 98.3/28.7/37.6 & 91.1/ \;1.8/ \;3.8 & 91.7/ \;2.2/ \;4.6 & 97.9/\textbf{60.6}/\textbf{59.6} \\
    \multicolumn{1}{c|}{} & \multicolumn{1}{c|}{Toothbrush} & 95.6/- & 88.9/- & 97.7/55.2/55.8 & \textbf{99.0}/53.6/58.8 & 98.4/34.9/45.7 & 76.9/ \;4.0/ \;7.7 & 93.7/20.4/ \;9.8 & \textbf{99.0}/\textbf{78.7}/\textbf{72.8} \\
    \multicolumn{1}{c|}{} & \multicolumn{1}{c|}{Transistor} & 92.3/- & 71.7/- & 64.5/23.6/15.1 & 85.9/42.3/45.2 & \textbf{97.9}/\textbf{59.5}/\textbf{64.6} & 53.2/ \;5.8/11.4 & 85.5/25.0/30.7 & 95.1/15.6/31.7 \\
    \multicolumn{1}{c|}{} & \multicolumn{1}{c|}{Zipper} & 94.8/- & 86.1/- &  98.3/\textbf{74.3}/\textbf{69.3} & \textbf{98.5}/53.9/60.3 & 96.8/40.1/49.9 & 67.4/ \;3.5/ \;7.6 & 66.9/ \;5.3/ \;7.4 & 96.2/60.7/60.0 \\
    \midrule
    \multicolumn{1}{c|}{\multirow{5}[1]{*}{\begin{turn}{-90}Textures\end{turn}}} & \multicolumn{1}{c|}{Carpet} & 97.6/- & 95.5/- & 98.6/\textbf{78.7}/\textbf{73.1} & 99.0/58.5/60.4 & 98.5/49.9/51.1 & 89.2/18.8/44.3 & \textbf{99.1}/70.6/66.0 & 98.6/42.2/46.4 \\
    \multicolumn{1}{c|}{} & \multicolumn{1}{c|}{Grid} & 71.0/- & 82.3/- & 98.7/44.5/46.2 & \textbf{99.2}/46.0/47.4 & 96.5/23.0/28.4 & 63.1/ \;0.7/ \;1.9 & 52.4/ \;1.1/ \;1.9 & 96.6/\textbf{66.0}/\textbf{64.1} \\
    \multicolumn{1}{c|}{} & \multicolumn{1}{c|}{Leather} & 84.8/- & 96.7/- & 97.3/\textbf{60.3}/57.4 & \textbf{99.3}/38.0/45.1 & 98.8/32.9/34.4 & 97.3/38.9/43.2 & 99.0/45.9/44.0 & 98.8/56.1/\textbf{62.3} \\
    \multicolumn{1}{c|}{} & \multicolumn{1}{c|}{Tile} & 80.5/- & 85.3/- &  \textbf{98.0}/\textbf{93.6}/\textbf{86.0} & 95.3/48.5/60.5 & 91.8/42.1/50.6 & 87.0/35.2/36.6 & 90.1/43.9/51.6 & 92.4/65.7/64.1 \\
    \multicolumn{1}{c|}{} & \multicolumn{1}{c|}{Wood} & 89.1/- & 80.5/- &  \textbf{96.0}/\textbf{81.4}/\textbf{74.6} & 95.3/47.8/51.0 & 93.2/37.2/41.5 & 84.7/30.9/37.3 & 92.3/44.1/46.6 & 93.3/43.3/43.5 \\
    \midrule
    \multicolumn{2}{c|}{Mean} & 89.5/- & 84.9/- & 87.2/52.5/48.6 & 96.1/48.6/53.8 & \textbf{96.8}/43.4/49.5 & 75.6/13.3/19.5 & 85.1/27.6/31.0 & \textbf{96.8}/\textbf{52.6}/\textbf{55.5} \\
    \bottomrule
    \end{tabular}
  }
    \caption{Comparison with SOTA methods on MVTec-AD dataset for multi-class anomaly localization with $AUROC_{seg}$/$AP_{seg}$/$F1max_{seg}$ metrics.}
  \label{tab:mvtecpx}%
\end{table*}%

\begin{table}[htbp]
  \centering
  \resizebox{0.8\linewidth}{!}{
    \begin{tabular}{c|cc|cc|c}
    \toprule
    \multirow{2}[4]{*}{Method} & \multicolumn{2}{c|}{Non-Diffusion } & \multicolumn{3}{c}{Diffusion-based} \\
\cmidrule{2-6}          & DRAEM & UniAD & DDPM  & LDM   & Ours \\
    \midrule
    PRO   & 71.1  & 90.4  & 49.0    & 66.3  & \textbf{90.7} \\
    \bottomrule
    \end{tabular}}
  \caption{Multi-class anomaly localization results with PRO metric on MVTec-AD datasets.}
  \label{tab:pro}%
\end{table}%

\noindent\textbf{MVTec-3D dataset.} MVTec-3D~\cite{Bergmann_2022} dataset comprises 4,147 scans obtained using a high-resolution industrial 3D sensor. It consists of 10 categories with both RGB images and 3D point clouds respectively. The training set contains 2,656 images with only anomaly-free samples. The test set consists of 1,197 images, including both normal and abnormal samples. Only RGB images are used in this experiment.

\noindent\textbf{Medical dataset.} We also merge three types of medical datasets  BraTS2021~\cite{baid2021rsna}, BTCV~\cite{landman2015miccai} and LiTs~\cite{bilic2023liver} into one \textit{Medical} dataset for multi-class anomaly detection. The training set contains 9,042 slices and the test set consists of 5,208 slices. 

\noindent\textbf{Evaluation Metrics.} Following prior works, Area Under the Receiver Operating Characteristic Curve (AUROC), Average Precision (AP) and F1-score-max (F1max) are used in both anomaly detection and anomaly localization, where $cls$ represents the image level anomaly detection and $seg$ represents the pixel level anomaly localization. Also, Per-Region-Overlap (PRO) is used in anomaly localization. The DICE score is commonly used in the medical field.

\subsection{Implementation Details}
All images in MVTec-AD and VisA are resized to 256 $\times$ 256. For the denoising network, we adopt the $4$-th block of SGDB for connection to SDDB. In this experiment, we adopt ResNet50 as the feature extraction network and choose $n\in \{2,3,4\}$ as the feature layers used in calculating the anomaly localization. We utilized the KL method as the Auto-encoder and fine-tune the model before training the denoising network. We train for 1000 epochs on a single NVIDIA Tesla V100 32GB with a batch size of 12. Adam optimiser~\cite{loshchilov2019decoupled} with a learning rate of 1e$^{-5}$ is set. A Gaussian filter with $\sigma=5$ is used to smooth the anomaly localization score. For anomaly detection, the anomaly score of the image is the maximum value of the averagely pooled anomaly localization score which undergoes 8 rounds of global average pooling operations with a size of 8 $\times$ 8. During inference, the initial denoising timestep $T$ is set from 1,000. We use DDIM~\cite{DBLP:conf/iclr/SongME21} as the sampler with 10 steps by default. 

\subsection{Comparison with SOTAs}
We conduct and analyze a range of qualitative and quantitative comparison experiments on MVTec-AD, VisA, MVTec-3D and \textit{Medical} datasets. We choose a synthesizing-based method DRAEM~\cite{zavrtanik2021draem}, three embedding-based methods MKD~\cite{salehi2021multiresolution}, PaDiM~\cite{defard2021padim} and RD4AD~\cite{deng2022anomaly}, a reconstruction-based method EdgRec~\cite{liu2022reconstruction}, a unified SOTA UniAD~\cite{NEURIPS2022_1d774c11} method and diffusion-based DDPM and LDM methods. Specifically, we categorize the aforementioned methods into two types: non-diffusion and diffusion-based methods. For the experiments on \textit{Medical} dataset, we follow the BMAD~\cite{bao2023bmad} benchmark and add two methods STFPM~\cite{yamada2021reconstruction} and CFLOW~\cite{gudovskiy2022cflow} for comparison.

\begin{table}[htbp]
  \centering
     \resizebox{\linewidth}{!}{
    \begin{tabular}{p{7em}<{\centering}|p{3em}<{\centering}p{3em}<{\centering}|p{3em}<{\centering}p{3em}<{\centering}|p{4em}<{\centering}}
    \toprule
    \multicolumn{1}{c|}{\multirow{2}[3]{*}{Metrics}} & \multicolumn{2}{c|}{Non-Diffusion} & \multicolumn{3}{c}{Diffusion-based} \\
    \cmidrule{2-6}    \multicolumn{1}{c|}{} & DRAEM & UniAD & DDPM  & LDM   & \textbf{Ours} \\
    \midrule
    $AUROC_{cls}$ & 63.2 & 78.9 & 66.3 & 68.5 & \textbf{84.6} \\
    $AP_{cls}$ & \multicolumn{1}{c}{86.1} & \multicolumn{1}{c|}{93.4} & \multicolumn{1}{c}{78.0} & \multicolumn{1}{c|}{90.6} & \multicolumn{1}{c}{\textbf{94.8}} \\
    $F1max_{cls}$ & \multicolumn{1}{c}{89.2} & \multicolumn{1}{c|}{91.4} & \multicolumn{1}{c}{86.6} & \multicolumn{1}{c|}{91.6} & \multicolumn{1}{c}{\textbf{95.5}} \\
    \midrule
    $AUROC_{seg}$ & \multicolumn{1}{c}{93.2} & \multicolumn{1}{c|}{\textbf{96.5}} & \multicolumn{1}{c}{90.7} & \multicolumn{1}{c|}{92.2} & 96.4 \\
    $AP_{seg}$ & \multicolumn{1}{c}{16.8} & \multicolumn{1}{c|}{21.2} &  \;\;6.0   & \;\;9.3     & \textbf{25.3} \\
    $F1max_{seg}$ & 20.2  & 28.0    & 10.7  & 13.5   & \textbf{32.2} \\
    $PRO$ & 55.0  & \textbf{88.1}  & 69.7  & 73.8    & 87.8 \\
    \bottomrule
    \end{tabular}}
    \caption{Quantitative comparisons on MVTec-3D dataset.}
  \label{tab:mvtec3d}%
\end{table}%

\begin{figure}[htbp]
\centering
\includegraphics[width=0.42\textwidth]{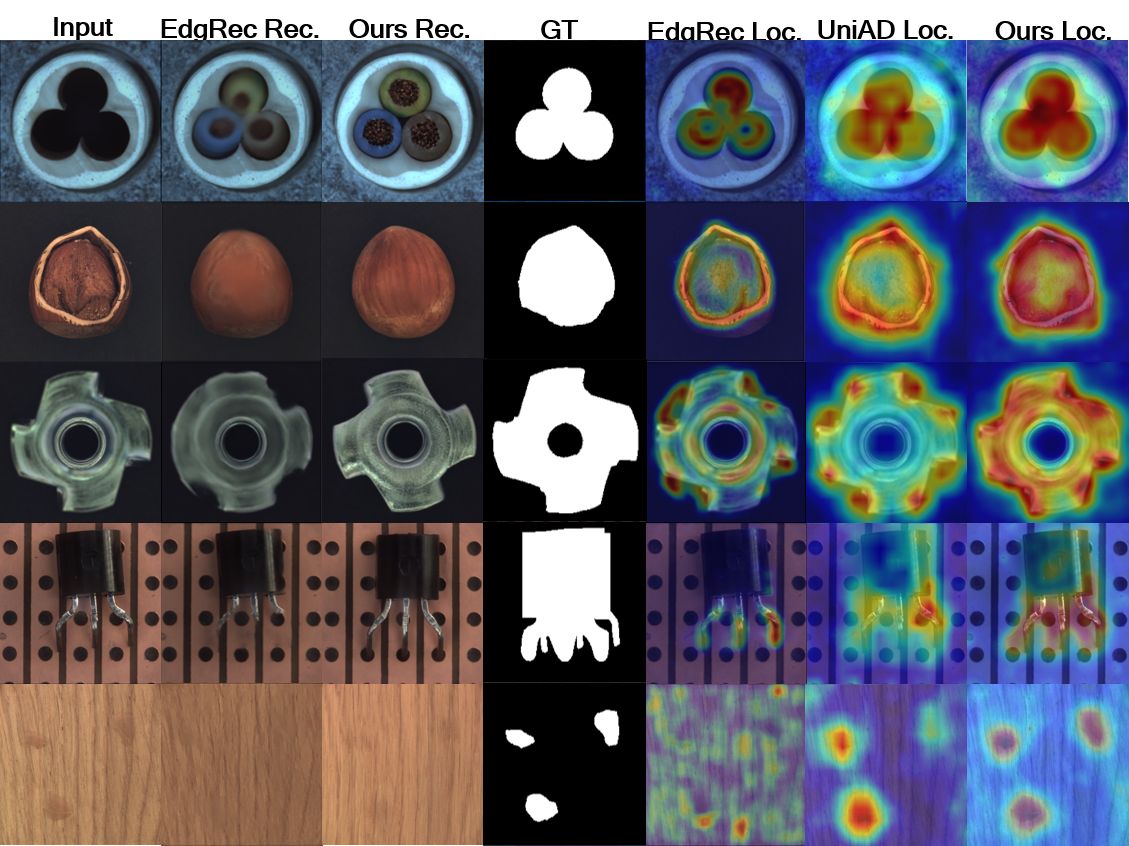} % Reduce the figure size so that it is slightly narrower than the column. Don't use precise values for figure width.This setup will avoid overfull boxes.
\caption{Qualitative illustration on MVTec-AD dataset.}
\label{fig:mvtec}
\end{figure}

\begin{figure}[t]
\centering
\includegraphics[width=0.43\textwidth]{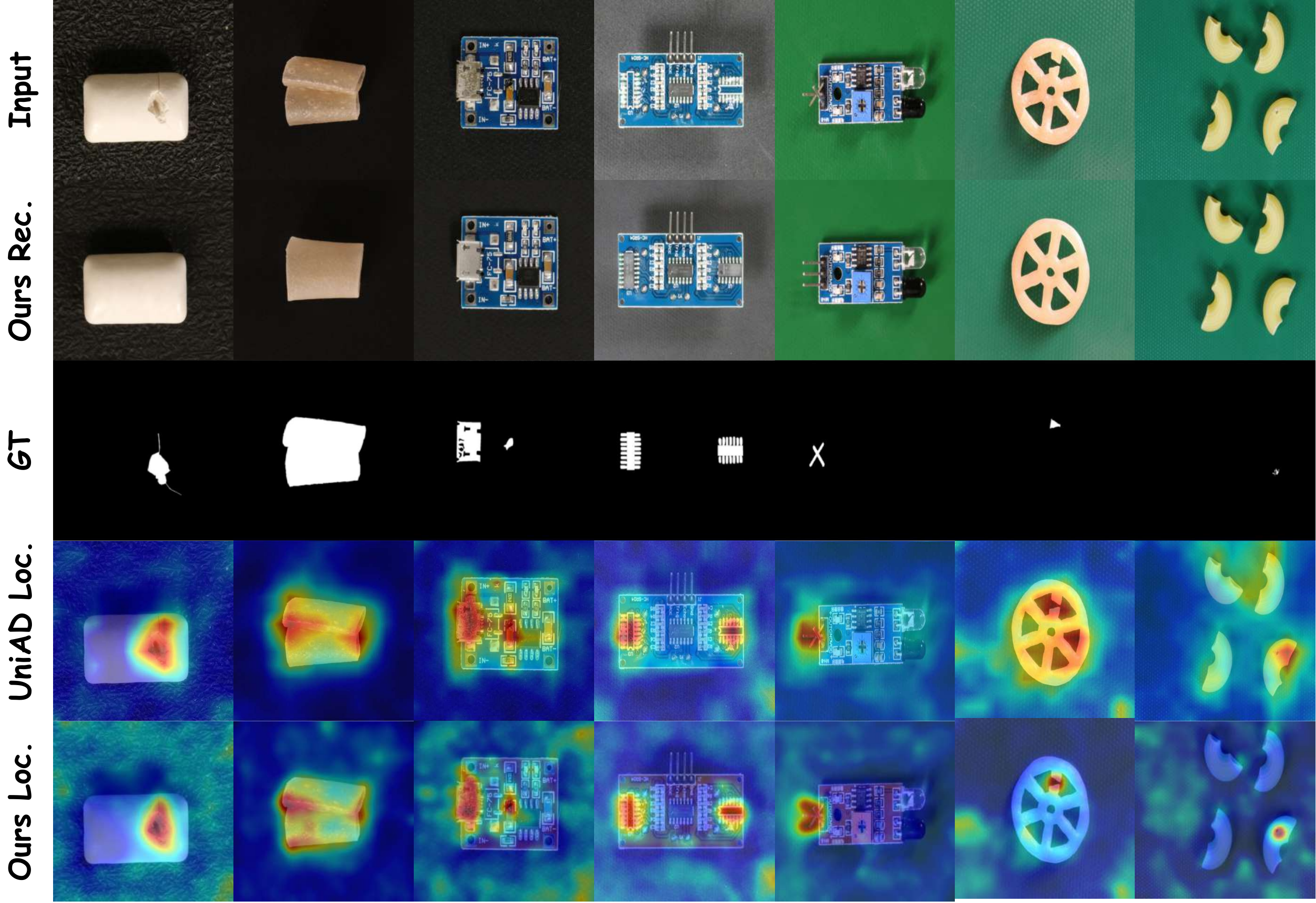} 
\caption{Qualitative results on VisA dataset.}
\label{fig:visa}
\end{figure}

\noindent\textbf{Qualitative Results.}
We conducted substantial qualitative experiments on MVTec-AD and VisA datasets to visually demonstrate the superiority of our method in image reconstruction and the accuracy of anomaly localization. As shown in Figure~\ref{fig:mvtec}, our method exhibits better reconstruction capabilities for anomalous regions compared to the EdgRec on MVTec-AD dataset. In comparison to UniAD shown in Figure~\ref{fig:visa}, our method exhibits more accurate anomaly localization abilities on VisA dataset. More qualitative results will be presented in \textit{Appendix}.

\noindent\textbf{Quantitative Results.} As shown in Table~\ref{tab:mvtecsp} and in Table~\ref{tab:mvtecpx}, our method achieves SOTA AUROC/AP/F1max metrics of 97.2/99.0/96.5 and 96.8/52.6/55.5 for image-wise and pixel-wise respectively for multi-class setting on MVTec-AD dataset. For the diffusion-based methods, our approach significantly outperforms existing DDPM and LDM methods in terms of 11.7$\uparrow$ in AUROC and 25$\uparrow$ in AP for anomaly localization. For non-diffusion methods, our approach surpasses existing methods in both metrics, especially at the pixel level, where our method exceeds UniAD by 9.2$\uparrow$/6.0$\uparrow$ in AP/F1max. Our method has also demonstrated its superiority on VisA dataset, as shown in Table~\ref{tab:visa}. Our approach exhibits significant improvements compared to diffusion-based methods of 30.1$\uparrow$/9.4$\uparrow$ than the LDM method in image/pixel AUROC. It also performs well compared to UniAD by 4.9$\uparrow$/6.0$\uparrow$ in pixel AP/F1max metrics. Detailed experiments for each category are provided in \textit{Appendix}. 
We have extended the method to 3D datasets and medical domain datasets. Table~\ref{tab:mvtec3d} and Table~\ref{tab:medical} show the effectiveness and scalability of our method on MVtec-3D and \textit{Medical} datasets, with results surpassing the state of the art (SOTA).

\begin{table}[htbp]
  \centering
    \resizebox{\linewidth}{!}{
    \begin{tabular}{p{4.25em}|ccccccc|c}
    \toprule
    \multicolumn{1}{c|}{Metrics} & MKD & CFLOW  & RD4AD & PaDiM & PatchCore & STFPM   & UniAD & Ours \\
    \midrule
    \multicolumn{1}{c|}{$AUROC_{cls}$} & 70.9 & 62.0 & 74.7  & 64.6  & 76.0    & 72.2  & 76.4  & \textbf{77.2} \\
    \multicolumn{1}{c|}{$AUROC_{seg}$} & 92.8 & 93.2 & 96.2  & 93.0    & 96.8  & 93.4  & 96.7  & \textbf{96.9} \\
    \multicolumn{1}{c|}{$PRO$}   & 79.3 & 79.0 & 88.0    & 79.2  & 86.6  & 86.0 & 87.4  & \textbf{87.7} \\
    \multicolumn{1}{c|}{$DICE$}  & 21.9 & 13.5 & 19.5  & 15.2  & 21.7  & 17.1 &  28.7 & \textbf{32.3} \\
    \bottomrule
    \end{tabular}}
    \caption{Quantitative comparisons on \textit{Medical} dataset.}
  \label{tab:medical}%
\end{table}%

\subsection{Ablation Studies}
\noindent\textbf{The architecture design of DiAD.}
We investigate the importance of each module in DiAD as shown in Table~\ref{tab:diad}. SD indicates only the diffusion model without connecting to the SG network which is the LDM's architecture. MSG indicates only the middle block of the SG network adding to the middle of SD. SGEB3 and SGEB4 indicate directly skip-connecting to the corresponding SDDB. When connecting SGDB3 and SGDB4 at the same time, more details of the original images are preserved in terms of texture, but the reconstruction ability for large anomaly areas decreases. Using the combination of IN+SiLU in the SFF block yields better results compared to using BN+ReLU.
\begin{table}[t]
  \centering
  \resizebox{1\linewidth}{!}{
    \begin{tabular}{cccccc|cc}
    \toprule
    SD    & MSG   & SGEB3 & SGEB4 & BN+ReLU & IN+SiLU   & $cls$  & $seg$ \\
    \midrule
    $\checkmark$     &       &       &       &       &       & 79.3  & 89.5 \\
    $\checkmark$     & $\checkmark$     &       &       &       &       & 95.1  & 91.1 \\
    $\checkmark$     & $\checkmark$     & $\checkmark$     &       &       &       & 95.3  & 89.1 \\
    $\checkmark$     & $\checkmark$     & $\checkmark$      & $\checkmark$     &      &        & 93.8  & 91.2 \\
    $\checkmark$     & $\checkmark$     & $\checkmark$     &        & $\checkmark$    &        & 96.7  & 96.7 \\
    $\checkmark$     & $\checkmark$     & $\checkmark$     &       &        & $\checkmark$     & \textbf{97.2}  & \textbf{96.8} \\
    \bottomrule
    \end{tabular}}
    \caption{Ablation studies on the design of DiAD with AUROC metrics.}
  \label{tab:diad}
\end{table}%

\noindent\textbf{Effect of pre-trained feature extractors.}
Table~\ref{tab:backbone} shows the quantitative comparison of using different pre-trained backbones as feature extraction networks. 
ResNet50 achieved the best performance in anomaly classification metrics, while WideResNet101 excelled in anomaly segmentation.
\begin{table}[htbp]
  \centering
  \resizebox{1\linewidth}{!}{
    \begin{tabular}{cc|ccccccc}
    \toprule
    \multicolumn{2}{c|}{Backbone} & $AUROC_{cls}$ & $AP_{cls}$ & $F1max_{cls}$ & $AUROC_{seg}$ & $AP_{seg}$ & $F1max_{seg}$ & $PRO$ \\
    \midrule
    \multirow{2}[2]{*}{VGG} & 16    & 91.8  & 97.2  & 93.9  & 92.1  & 47.2  & 50.5  & 80.1 \\
          & 19    & 91.3  & 96.9  & 93.7  & 92.3  & 47.5  & 50.6  & 80.4 \\
    \midrule
    \multirow{4}[2]{*}{ResNet} & 18    & 94.7  & 98.1  & 96.0    & 96.0    & 49.9  & 53.3  & 89.1 \\
          & 34    & 95.2  & 98.3  & 95.7  & 96.2  & 51.2  & 54.5  & 89.6 \\
          & 50    & \textbf{97.2} & \textbf{99} & \textbf{96.5} & 96.8  & 52.6  & 55.5  & 90.7 \\
          & 101   & 96.2  & 98.4  & \textbf{96.5} & \textbf{96.9} & 52.9  & 56.4  & 91.2 \\
    \midrule
    \multirow{2}[2]{*}{WideResNet} & 50    & 95.9  & 98.6  & \textbf{96.5} & 96.4  & 51.8  & 55.1  & 89.3 \\
          & 101   & 95.6  & 98.3  & 95.8  & \textbf{96.9} & \textbf{54.6} & \textbf{56.5} & \textbf{91.4} \\
    \midrule
    \multirow{3}[2]{*}{EfficientNet} & b0    & 93.5  & 97.7  & 94.7  & 94.0    & 50.0    & 52.4  & 84.0 \\
          & b2    & 94.2  & 98.0    & 95.1  & 94.1  & 48.6  & 52.1  & 84.2 \\
          & b4    & 92.8  & 97.5  & 94.8  & 93.6  & 47.2  & 50.7  & 83.5 \\
    \bottomrule
    \end{tabular}}
    \caption{Ablation studies on different feature extractors.}
  \label{tab:backbone}%
\end{table}%

\noindent\textbf{Effect of feature layers used in anomaly score calculating.} After extracting feature maps of 5 different scales using a pre-trained backbone, the anomaly scores are calculated by computing the cosine similarity between feature maps from different layers. The experimental results, as shown in Table~\ref{tab:feature}, indicate that using feature maps from layers $f_2$, $f_3$, and $f_4$ (with corresponding sizes of $64\times64$, $32\times32$, and $16\times16$) yields the best performance.
\begin{table}[htbp]
  \centering
  \resizebox{1\linewidth}{!}{
    \begin{tabular}{ccccc|cccccc}
    \toprule
    $f_1$     & $f_2$     & $f_3$     & $f_4$     & $f_5$     & $AUROC_{cls}$ & $AP_{cls}$ & $F1max_{cls}$ & $AUROC_{seg}$ & $AP_{seg}$ & $F1max_{seg}$ \\
    \midrule
    \checkmark     & \checkmark     & \checkmark     & \checkmark     & \checkmark    & 93.8  & 97.8  & 95.0    & 94.0    & 42.0    & 45.9 \\
    \checkmark     & \checkmark     & \checkmark     & \checkmark     &       & 96.7  & 98.7  & 96.1  & 96.7  & 52.5  & 55.2 \\
          & \checkmark     & \checkmark     & \checkmark     & \checkmark     & 93.4  & 97.1  & 93.6  & 95.2  & 48.5  & 51.3 \\
     \checkmark     & \checkmark     & \checkmark     &       &       & 97.1  & \textbf{99.0} & \textbf{96.8} & 96.4  & 49.4  & 53.1 \\
          & \checkmark     & \checkmark     & \checkmark     &       & \textbf{97.2} & \textbf{99.0} & 96.5  & \textbf{96.8} & \textbf{52.6} & \textbf{55.5} \\
          & \checkmark     & \checkmark     &       &       & 94    & 97.4  & 94.2  & 95.3  & 48.5  & 51.7 \\
          &       & \checkmark     & \checkmark    &       & 97.1  & \textbf{99.0}     & \textbf{96.8}     & 96.4  & 49.4     & 53.1 \\
    \bottomrule
    \end{tabular}}
    \caption{Ablation studies on the feature layers used in calculating the anomaly localization score based on ResNet50.}
  \label{tab:feature}%
\end{table}%

\begin{figure}[htbp]
\centering
\includegraphics[width=0.37\textwidth]{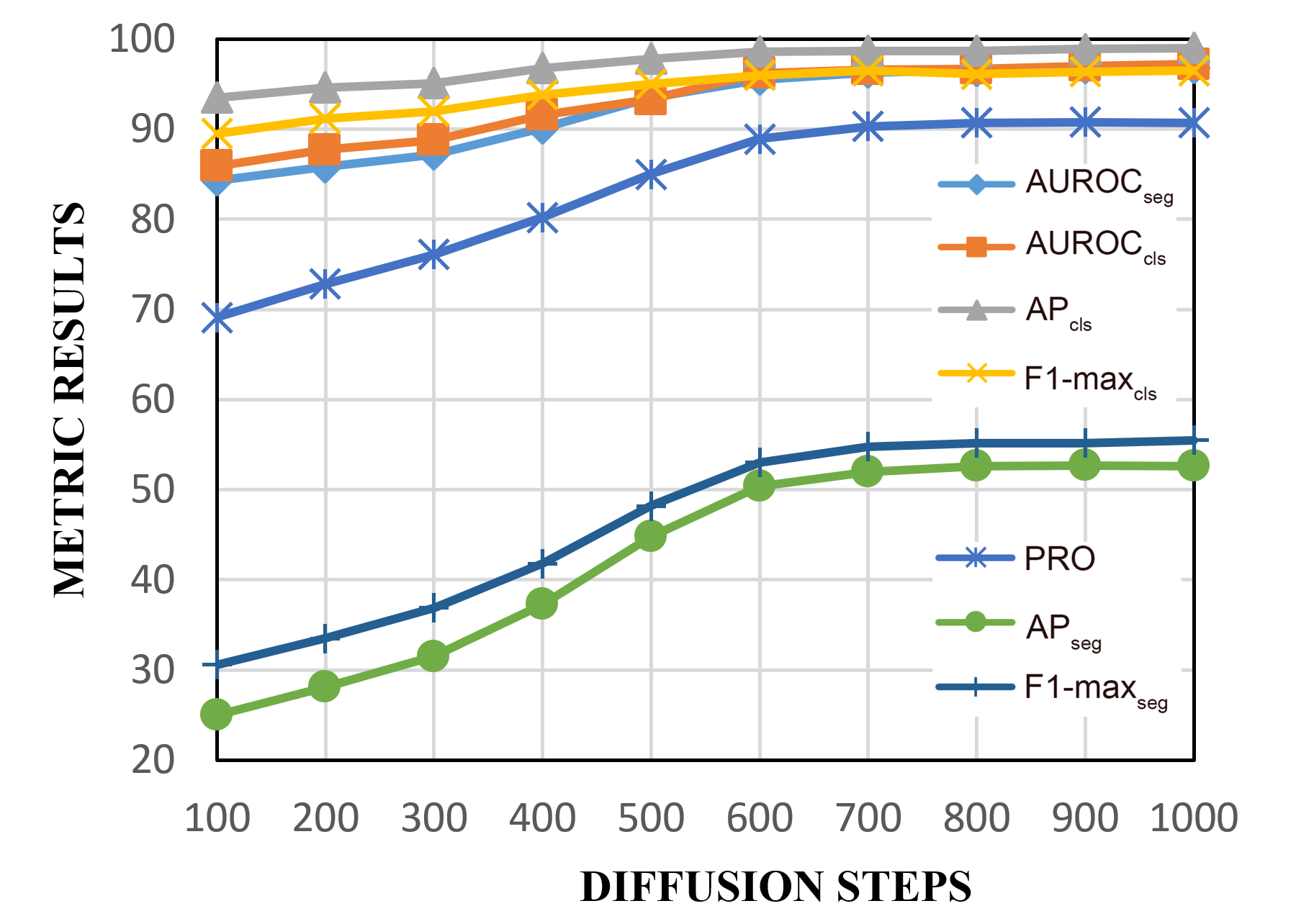} 
\caption{Ablation studies on different diffusion timesteps.}
\label{fig:difstep}
\end{figure}

\noindent\textbf{Effect of forward diffusion timesteps.} Increasing the number of diffusion steps in the forward process impacts the performance of image reconstruction. The experimental results, depicted in Figure~\ref{fig:difstep}, indicate that with an increasing number of forward diffusion steps, the image approaches pure Gaussian noise, while the anomaly reconstruction ability improves as well. Nevertheless, when the number of forward diffusion steps is less than 600, a significant decline in performance occurs because the number of steps is insufficient for anomaly reconstruction.

\section{Conclusion}
This paper proposes a diffusion-based DiAD framework to address the issue of category and semantic loss in the stable diffusion model for multi-class anomaly detection. We propose the Semantic-Guided network and Spatial-aware Feature Fusion block to better reconstruct the abnormal regions while maintaining the same semantic information as the input image. Our approach achieves state-of-the-art performance on MVTec-AD and VisA datasets, significantly outperforming the non-diffusion and diffusion-based methods. 

\noindent\textbf{Limitation.}  Although our method has demonstrated exceptional performance in reconstructing anomalies, it can be susceptible to the influence of background impurities, resulting in errors in localization and classification. In the future, we will further explore diffusion models and enhance the background's anti-interference capability for multi-class anomaly detection. Additionally, we will incorporate multi-modal assistance in our anomaly detection. Lastly, we will utilize larger models to enhance reconstruction performance.

\bibliography{aaai24wo}

\appendix
\section{Appendices}
\subsection{Effect of DDIM sampler steps}
In order to accelerate the sampling speed in the denoising process, UiAD adopts the DDIM sampling strategy. We investigated the impact of different DDIM sampler steps on the results, as shown in Table \ref{tab:ddim}. The results indicate that increasing the number of sampling steps does not significantly affect the results. Therefore, using a 10-step sampling process can achieve the best performance while greatly accelerating the sampling speed.

\begin{table}[htbp]
  \centering
  \resizebox{1\linewidth}{!}{
    \begin{tabular}{c|ccccccc}
    \toprule
    Steps & 1     & 5     & 10    & 20    & 50    & 100   & 200 \\
    \midrule
    $seg$  & 72.5  & 96.5  & \textbf{96.8} & \textbf{96.8} & 96.7  & 96.7  & \textbf{96.8} \\
    $cls$  & 66.1  & 96.4  & \textbf{97.2} & 97.1  & 97.0    & 96.8  & 96.9 \\
    \bottomrule
    \end{tabular}}
    \caption{Ablation studies on DDIM sampler steps.}
  \label{tab:ddim}%
\end{table}%

\subsection{Effect of Global average pooling}
Global average pooling is used to reduce the potential occurrence of false positives. For \textit{m-n} in the table below, \textit{m} represents the iterations and \textit{n} represents the kernel size. Through quantitative analysis, the most effective approach is employing an $8\times8$ size global average pooling with \textit{8} iterations. Also, the best-performing combinations exhibit the same feature map size.

\noindent
\resizebox{\linewidth}{!}{
\begin{tabular}{l|llllllll}
\toprule
Global Average Pooling & 1-16 & 4-16 & 5-12 & 6-10 & 8-8  & 10-7 & 15-5 & 20-4 \\
\midrule
AUROC-cls & 96.0  & 96.7  & 96.9  & 97.1  & \textbf{97.2}  & \textbf{97.2}  & 97.0  & 96.8 \\
\bottomrule
\end{tabular}}

\begin{figure*}[htbp]
\centering
\includegraphics[width=0.7\textwidth]{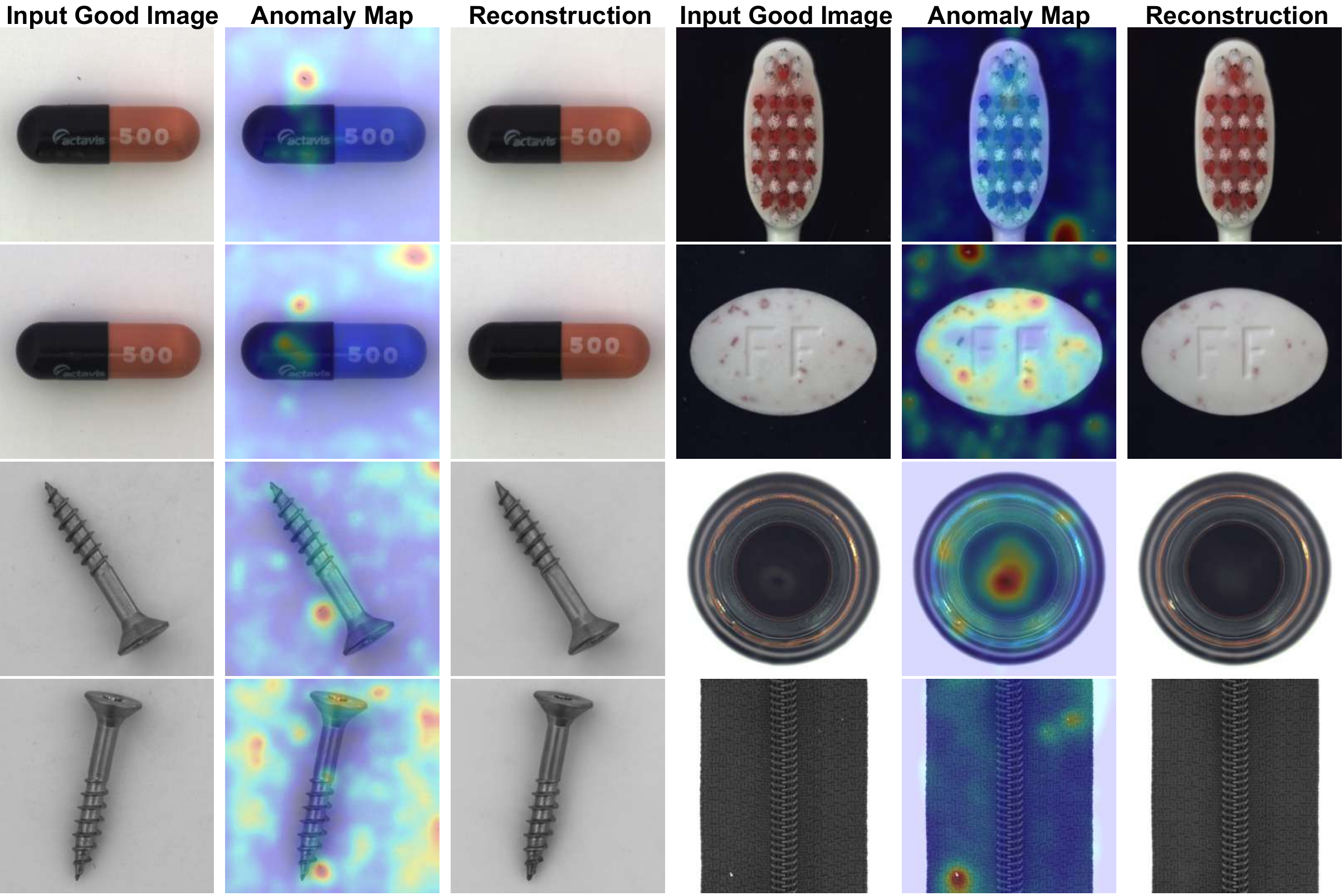} 
\caption{Visualization of false positive classifications and localizations.}
\label{limit}
\end{figure*}

\subsection{Limitations of the datasets}
 We found that there are several categories of image-level anomaly detection results that are significantly lower than others, such as capsules and screws. As shown in Fig \ref{limit}, we discovered some false positives in input good images during the test. Our method performs well in reconstructing the objects in the objects' main bodies, but the background region of the original image contains impurities, causing the pre-trained feature extraction network to extract features that perceive the background impurities as anomalies. As anomaly detection is expected to identify anomalies within the object rather than the background region, there are certain deficiencies in the Mvtec-AD as well as the VisA datasets that lead to false positives. In response to this issue, we increase the number of global average pooling operations to alleviate the problem of high anomaly scores caused by impurities in the background.

\subsection{Hyperparameters of DiAD}
We provided a comprehensive set of hyperparameters for the three models in DiAD as shown in Table \ref{tab:hyperparameters}.

\begin{table*}[htbp]
  \centering
    \begin{tabular}{l|ccc}
    \toprule
    \multicolumn{1}{c|}{\multirow{2}[4]{*}{Parameters Name}} & \multicolumn{3}{c}{Model Name} \\
\cmidrule{2-4}          & \multicolumn{1}{c|}{SD Denoising Network} & \multicolumn{1}{c|}{SG Network} & Autoencoder \\
    \midrule
    $z$ shape & \multicolumn{3}{c}{$32\times32\times4$} \\
    $|z|$  & \multicolumn{3}{c}{4096} \\
    Diffusion steps $T$ & \multicolumn{3}{c}{1000} \\
    DDIM sampling steps $T$ & \multicolumn{3}{c}{10} \\
    Noise Schedule & \multicolumn{3}{c}{linear} \\
    \midrule
    Model input shape & \multicolumn{1}{c|}{$32\times32\times4$} & \multicolumn{1}{c|}{$256\times256\times3$} & $256\times256\times3$ \\
    N params & \multicolumn{1}{c|}{859M} & \multicolumn{1}{c|}{471M} & 83.7M \\
    Embed dim & \multicolumn{1}{c|}{-} & \multicolumn{1}{c|}{-} & 4 \\
    Channels & \multicolumn{1}{c|}{320} & \multicolumn{1}{c|}{320} & 128 \\
    Num res blocks & \multicolumn{1}{c|}{2} & \multicolumn{1}{c|}{2} & 2 \\
    Channel Multiplier & \multicolumn{1}{c|}{1,2,4,4} & \multicolumn{1}{c|}{1,2,4,4} & 1,2,4,4 \\
    Attention resolutions & \multicolumn{1}{c|}{4,2,1} & \multicolumn{1}{c|}{4,2,1} & - \\
    Num Heads & \multicolumn{1}{c|}{8} & \multicolumn{1}{c|}{8} & - \\
    \midrule
    Batch Size & \multicolumn{3}{c}{12} \\
    Accumulate\_grad\_batches & \multicolumn{3}{c}{4} \\
    Epochs & \multicolumn{3}{c}{1000} \\
    Learning Rate & \multicolumn{3}{c}{1.0e-5} \\
    \bottomrule
    \end{tabular}%
    \caption{Hyperparameters for the DiAD. All models trained on a single NVIDIA Tesla V100 32GB.}
  \label{tab:hyperparameters}%
\end{table*}%

\begin{table*}[htbp]
  \centering
  \resizebox{1.0\linewidth}{!}{
    \begin{tabular}{c|cc|cc|c}
    \toprule
    \multicolumn{1}{c|}{\multirow{2}[3]{*}{Category}} & \multicolumn{2}{c|}{Non-Diffusion Method} & \multicolumn{3}{c}{Diffusion-based Method} \\
\cmidrule{2-6}    \multicolumn{1}{c|}{} & \multicolumn{1}{c}{DRAEM} & \multicolumn{1}{c|}{UniAD} & DDPM  & LDM   & Ours \\
    \midrule
    pcb1  & 71.9/72.2/70.0 & 92.8/92.7/87.8 & 54.1/47.7/67.1 & 51.2/46.9/66.8 & 88.1/88.7/80.7 \\
    pcb2  & 78.4/78.2/76.2 & 87.8/87.7/83.1 & 50.8/48.5/66.6 & 57.0/63.4/67.5 & 91.4/91.4/84.7 \\
    pcb3  & 76.6/77.4/74.7 & 78.6/78.6/76.1 & 53.4/51.2/66.8 & 62.7/69.6/72.0 & 86.2/87.6/77.6 \\
    pcb4  & 97.3/97.5/93.5 & 98.8/98.8/94.3 & 56.0/48.4/66.4 & 54.4/47.1/66.8 & 99.6/99.5/97.0 \\
    \midrule
    macaroni1 & 69.8/68.5/70.9 & 79.9/79.8/72.7 & 50.9/55.1/68.0 & 56.2/49.6/68.4 & 85.7/85.2/78.8 \\
    macaroni2 & 59.4/60.7/68.0 & 71.6/71.6/69.9 & 54.4/51.8/67.1 & 56.8/52.7/66.6 & 62.5/57.4/69.6 \\
    capsules & 83.4/91.1/82.1 & 55.6/55.6/76.9 & 58.9/62.7/78.2 & 57.7/71.4/77.3 & 58.2/69.0/78.5 \\
    candle & 69.3/73.9/68.0 & 94.1/94.0/86.1 & 52.7/48.3/66.6 & 50.4/52.2/68.2 & 92.8/92.0/87.6 \\
    \midrule
    cashew & 81.7/89.7/87.3 & 92.8/92.8/91.4 & 63.5/78.9/80.6 & 61.1/71.0/80.0 & 91.5/95.7/89.7 \\
    chewinggum & 93.7/97.1/91.0 & 96.3/96.2/95.2 & 50.9/65.6/80.0 & 53.9/65.8/81.3 & 99.1/99.5/95.9 \\
    fryum & 89.1/95.0/86.6 & 83.0/83.0/85.0 & 51.0/62.4/80.0 & 63.5/71.6/81.6 & 89.8/95.0/87.2 \\
    pipe\_fryum & 82.8/91.2/83.9 & 94.7/94.7/93.9 & 56.9/74.9/80.0 & 56.1/75.5/80.3 & 96.2/98.1/93.7 \\
    \midrule
    Mean  & 79.1/81.9/78.9 & 85.5/85.5/84.4 & 54.5/57.9/72.3 & 56.7/61.4/73.1 & 86.8/88.3/85.1 \\
    \bottomrule
    \end{tabular}}
  \caption{Comparison with SOTA methods on VisA dataset for multi-class anomaly detection with $AUROC_{cls}$/$AP_{cls}$/$F1max_{cls}$ metrics.}
  \label{tab:app_visa_det}%
\end{table*}%

\begin{table*}[htbp]
  \centering
  \resizebox{1.0\linewidth}{!}{
    \begin{tabular}{c|cc|cc|c}
    \toprule
    \multicolumn{1}{c|}{\multirow{2}[3]{*}{Category}} & \multicolumn{2}{c|}{Non-Diffusion Method} & \multicolumn{3}{c}{Diffusion-based Method} \\
\cmidrule{2-6}    \multicolumn{1}{c|}{} & \multicolumn{1}{c}{DRAEM} & \multicolumn{1}{c|}{UniAD} & DDPM  & LDM   & Ours \\
    \midrule
    pcb1  & 94.6/31.8/37.2/52.8 & 93.3/ \;3.9/ \;8.3/64.1 & 75.7/ \;1.1/ \;2.8/36.1 & 84.5/ \;2.1/ \;4.9/54.3 & 98.7/49.6/52.8/80.2 \\
    pcb2  & 92.3/10.0/18.6/66.2 & 93.9/ \;4.2/ \;9.2/66.9 & 76.2/ \;0.7/ \;1.6/30.8 & 89.5/ \;2.5/ \;6.7/52.7 & 95.2/ \;7.5/16.7/67.0 \\
    pcb3  & 90.8/14.1/24.4/42.9 & 97.3/13.8/21.9/70.6 & 83.3/ \;1.0/ \;2.5/56.1 & 94.4/ \;9.2/17.4/67.8 & 96.7/ \;8.0/18.8/68.9 \\
    pcb4  & 94.4/31.0/37.6/75.7 & 94.9/14.7/22.9/72.3 & 73.0/ \;1.4/ \;3.5/29.9 & 80.4/ \;2.1/ \;4.2/40.3 & 97.0/17.6/27.2/85.0 \\
    \midrule
    macaroni1 & 95.0/19.1/24.1/67.0 & 97.4/ \;3.7/ \;9.7/84.0 & 87.4/ \;0.4/ \;1.0/61.2 & 81.6/ \;0.3/ \;1.3/47.3 & 94.1/10.2/16.7/68.5 \\
    macaroni2 & 94.6/ \;3.9/12.4/65.2 & 95.2/ \;0.9/ \;4.3/76.6 & 84.8/ \;0.2/ \;0.6/54.1 & 87.2/ \;0.3/ \;0.6/57.2 & 93.6/ \;0.9/ \;2.8/73.1 \\
    capsules & 97.1/27.8/33.7/62.8 & 88.7/ \;3.0/ \;7.4/43.7 & 77.1/ \;1.1/ \;2.8/34.6 & 75.5/ \;1.1/ \;2.7/34.8 & 97.3/10.0/21.0/77.9 \\
    candle & 82.2/10.1/19.0/65.6 & 98.5/17.6/27.9/91.6 & 76.4/ \;0.4/ \;1.4/34.1 & 85.3/ \;0.9/ \;1.9/46.8 & 97.3/12.8/22.8/89.4 \\
    \midrule
    cashew & 80.7/ \;9.9/15.7/38.5 & 98.6/51.7/58.3/87.9 & 74.5/ \;2.7/ \;5.2/58.7 & 90.5/ \;5.1/10.1/68.3 & 90.9/53.1/60.9/61.8 \\
    chewinggum & 91.0/62.3/63.3/40.9 & 98.8/54.9/56.1/81.3 & 74.7/ \;1.4/ \;2.8/37.9 & 84.1/ \;3.1/ \;6.9/52.9 & 94.7/11.9/25.8/59.5 \\
    fryum & 92.4/38.8/38.5/69.5 & 95.9/34.0/40.6/76.2 & 85.7/ \;9.4/17.2/58.4 & 89.9/14.8/24.8/60.1 & 97.6/58.6/60.1/81.3 \\
    pipe\_fryum & 91.1/38.1/39.6/61.8 & 98.9/50.2/57.7/91.5 & 87.0/ \;6.9/12.9/69.6 & 96.4/31.0/37.2/77.6 & 99.4/72.7/69.9/89.9 \\
    \midrule
    Mean  & 91.3/23.5/29.5/58.8 & 95.9/21.0/27.0/75.6 & 79.7/ \;2.2/ \;4.5/46.8 & 86.6/ \;6.0/ \;9.9/55.0 & 96.0/26.1/33.0/75.2 \\
    \bottomrule
    \end{tabular}}
  \caption{Comparison with SOTA methods on VisA dataset for multi-class anomaly localization with $AUROC_{seg}$/$AP_{seg}$/$F1max_{seg}$/$PRO$ metrics.}
  \label{tab:app_visa_loc}%
\end{table*}%

\begin{figure*}[t]
\centering
\includegraphics[scale=0.16]{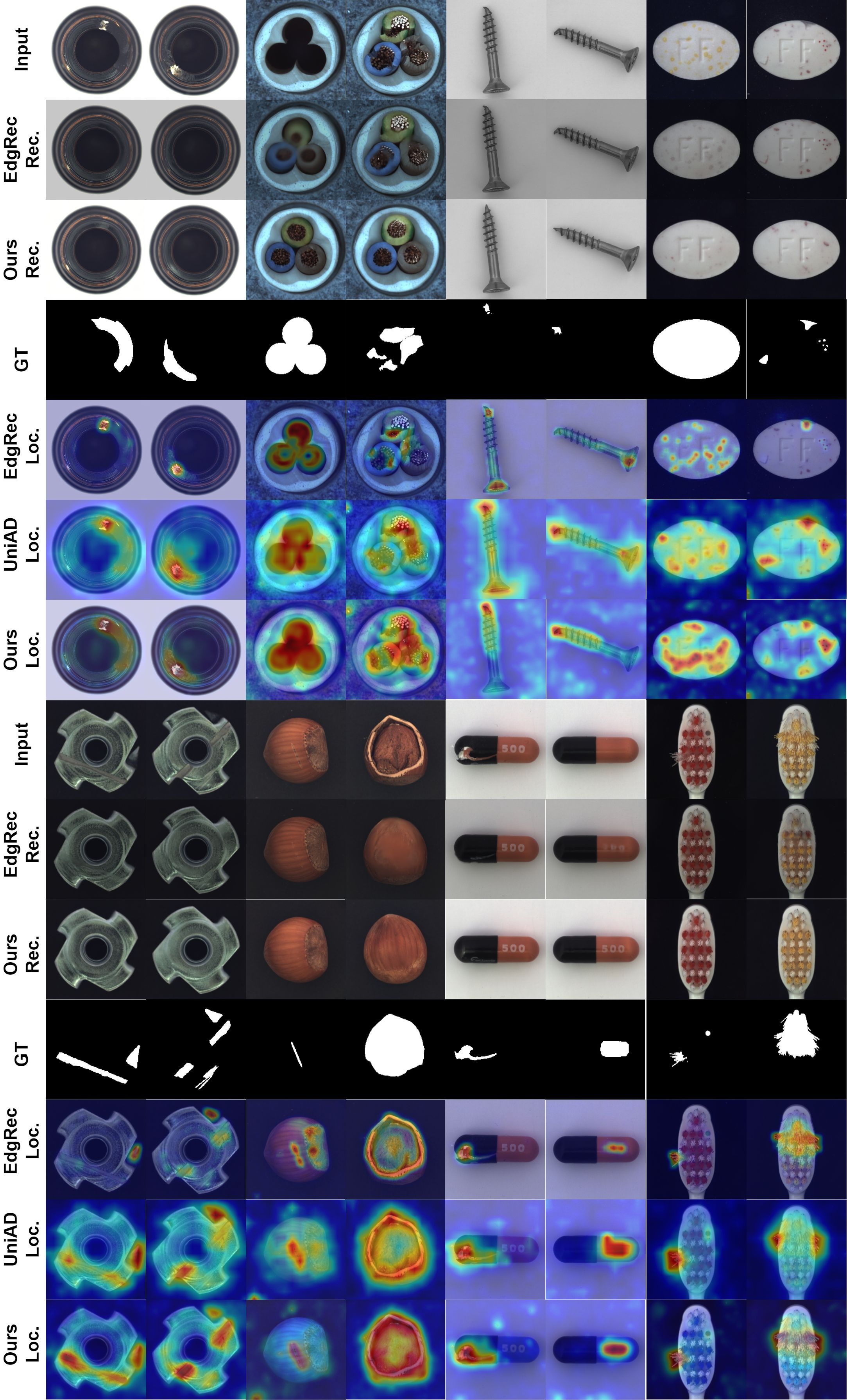} 
\caption{Qualitative comparison results for anomaly localization on MVTec-AD dataset.}
\label{fig:appendixmvtec}
\end{figure*}

\begin{figure*}[t]
\centering
\includegraphics[scale=0.18]{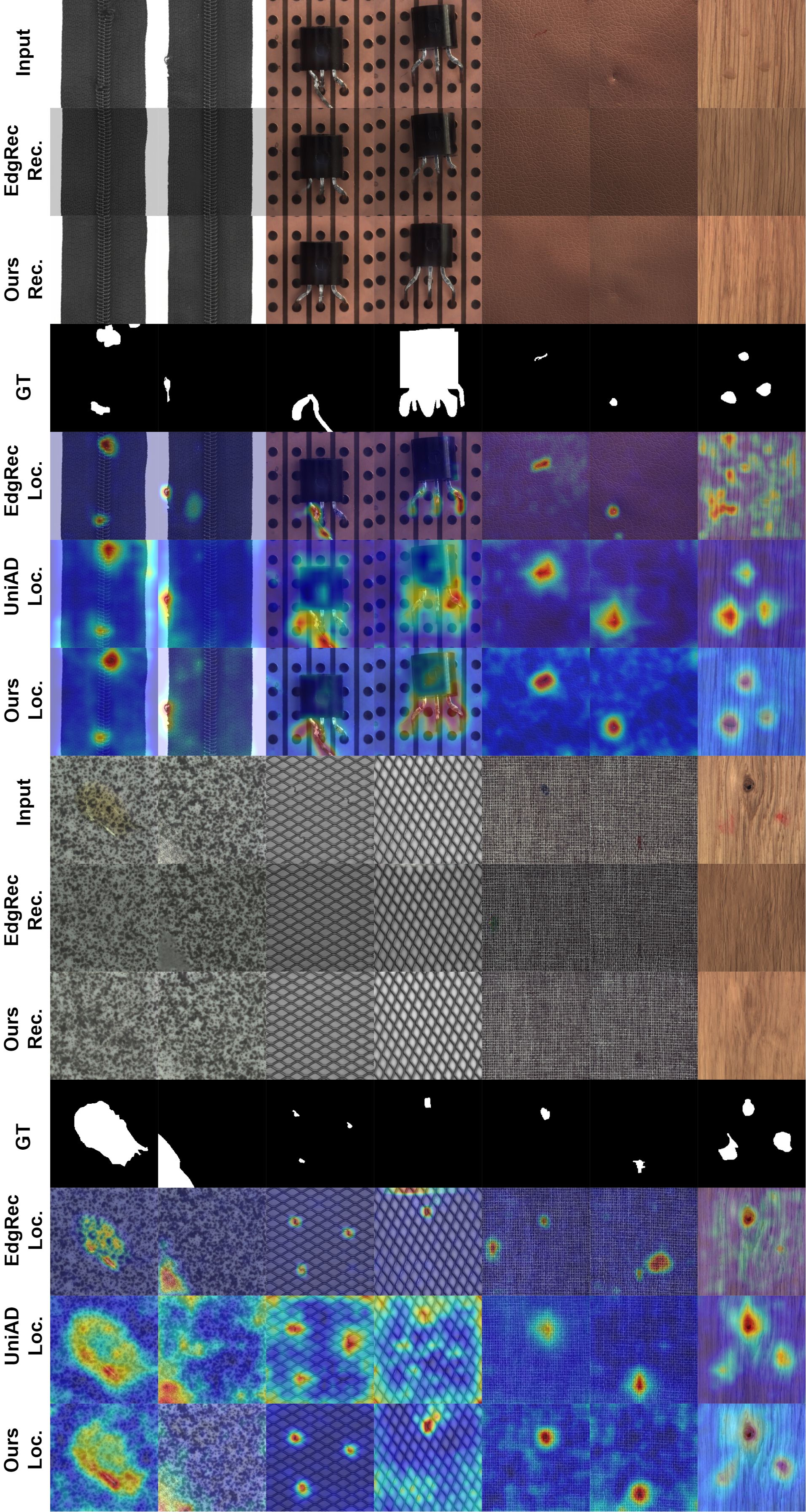} 
\caption{Qualitative comparison results for anomaly localization on MVTec-AD dataset.}
\label{fig:appendixmvtec2}
\end{figure*}

\begin{figure*}[t]
\centering
\includegraphics[width=0.9\textwidth]{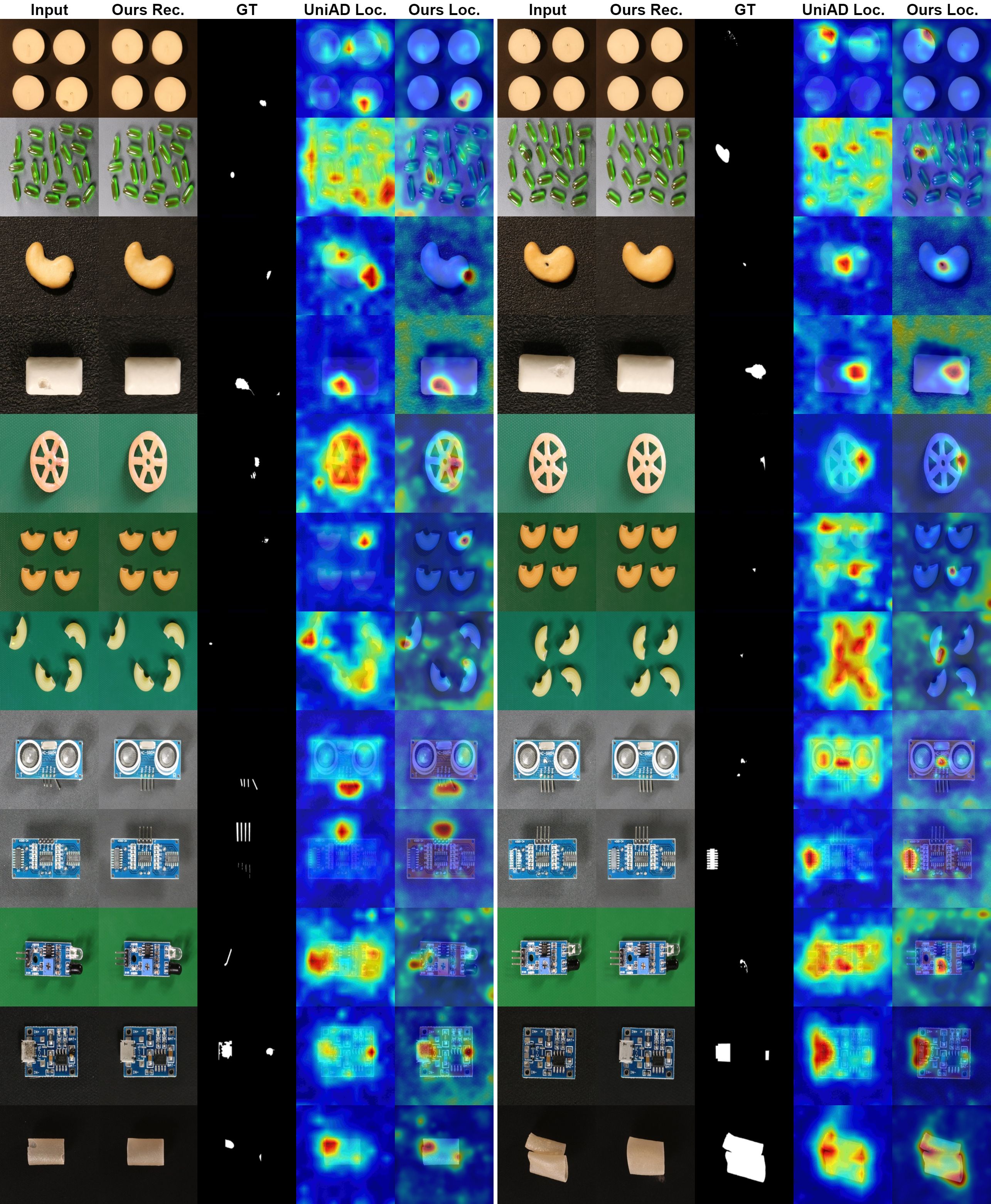} 
\caption{Qualitative comparison results for anomaly localization on VisA dataset.}
\label{fig:appendixvisa}
\end{figure*}

\end{document}